\documentclass[final,5p,times,twocolumn,authoryear]{elsarticle}
\usepackage{graphicx}
\usepackage{float}
\usepackage{caption}
\usepackage{subcaption}
\usepackage{tikz}
\usepackage{siunitx}
\usepackage{amsmath}
\usepackage{amssymb}
\usepackage{amsthm}
\usepackage[ruled,vlined]{algorithm2e}
\usepackage{appendix}
\usepackage[utf8x]{inputenc}
\usepackage[noend]{algpseudocode}
\usepackage{lineno}
\usepackage{adjustbox}
\usepackage{parskip} 
\usepackage[version=4]{mhchem}
\usepackage{tikz}
\usepackage{pgfplots}
\pgfplotsset{compat=1.17}
\usepackage{xcolor}

\usepackage[nolist,nohyperlinks]{acronym}
\acrodef{NN}{neural network}
\acrodef{PGNN}{physics-guided neural network}
\acrodef{PINN}{physics-informed neural network}
\acrodef{PBM}{physics-based models}
\acrodef{DDM}{Data-driven modeling}
\acrodef{HAM}{hybrid analysis and modeling}
\acrodef{ML}{machine learning}
\acrodef{MSE}{mean squared error}
\acrodef{RFMSE}{rolling forecast mean squared error}
\acrodef{MAV}{miniature air vehicle}
\acrodef{PDE}{partial differential equation}
\acrodef{ODE}{ordinary differential equations}
\acrodef{RNN}{recurrent neural network}
\acrodef{ROM}{reduced order model}
\acrodef{DOF}{degrees of freedom}
\acrodef{CG}{centre of gravity}
\acrodef{PCR}{principal component regression}
\acrodef{PCA}{principal component analysis}
\acrodef{NED}{North-East-Down}
\acrodef{PID}{Proportional-Integral-Derivative}
\acrodef{NN}{neural network}
\acrodef{PWA}{piecewise affine}
\acrodef{MILP}{mixed integer linear programming problem}
\acrodef{SMT}{satisfiability modulo theory}
\acrodef{IVP}{initial value problem}
\acrodef{MPC}{model predictive control}
\acrodef{PWL}{piecewise linear}
\acrodef{CFD}{computational fluid dynamics}
\acrodef{EKF}{Extended Kalman Filter}
\acrodef{RK4}{Runge-Kutta 4}
\acrodef{CoSTA}{Corrective source term approach}
\acrodef{APRBS}{Amplitude-modulated Pseudo-Random Binary Signal}
\acrodef{SciML}{Scientific Machine Learning}
\acrodef{ACD}{Anode-Cathode Distance}
\acrodef{UAV}{unmanned aerial vehicle}
\acrodef{PE}{persistency of excitation}

\theoremstyle{definition}

\definecolor{Tblue}{HTML}{1f77b4}
\definecolor{Torange}{HTML}{ff7f0e}
\definecolor{Tgreen}{HTML}{2ca02c}
\definecolor{Tred}{HTML}{d62728}
\definecolor{Tpurple}{HTML}{9467bd}
\definecolor{Tbrown}{HTML}{8c564b}

\makeatletter
\newenvironment{customlegend}[1][]{%
    \begingroup
    \pgfplots@init@cleared@structures
    \pgfplotsset{#1}%
}{
    \pgfplots@createlegend
    \endgroup
}

\def\addlegendimage{\pgfplots@addlegendimage}

\newcommand{\todo}[1]{}

\newcommand{\state}[0]{\mathbf{x}}
\newcommand{\xinput}[0]{\mathbf{u}}
\newcommand{\source}[0]{\mathbf{f}}

\newcommand{\residual}[0]{\boldsymbol{\sigma}}

\newcommand{\differentiate}[0]{{\mathcal{L}\mspace{2mu}}}

\newcommand{\net}[0]{\hat{\mathbf{f}}}
\newcommand{\netinput}[0]{\mathbf{z}}
\newcommand{\netoutput}[0]{\mathbf{y}}
\newcommand{\netparameters}[0]{\boldsymbol{\theta}}
\newcommand{\weights}[0]{\mathbf{w}}
\newcommand{\WEIGHTS}[0]{\mathbf{W}}
\newcommand{\netweights}[1]{\WEIGHTS^{#1}}
\newcommand{\netweight}[2]{\weights^{#1}_{#2}}
\newcommand{\netbiases}[1]{\mathbf{b}^{#1}}
\newcommand{\netbias}[2]{{b}^{#1}_{#2}}
\newcommand{\totalloss}[0]{{C}}
\newcommand{\loss}[0]{{L}}
\newcommand{\wpenalty}[0]{{R}}

\allowdisplaybreaks

\begin{document} 
\begin{frontmatter}
\title{A novel corrective-source term approach to modeling unknown physics in aluminum extraction process}

\author[NTNU]{Haakon Robinson\corref{mycorrespondingauthor}}
\ead{haakon.robinson@ntnu.no}

\author[NTNU]{Erlend Lundby\corref{mycorrespondingauthor}}
\ead{erlend.t.b.lundby@ntnu.no}
\cortext[mycorrespondingauthor]{Equal contributions}

\author[NTNU]{Adil Rasheed}
\ead{adil.rasheed@ntnu.no}

\author[NTNU]{Jan Tommy Gravdahl}
\ead{jan.tommy.gravdahl@ntnu.no}

\address[NTNU]{Department of Engineering Cybernetics, Norwegian University of Science and Technology, O. S. Bragstads plass 2, Trondheim, NO-7034, Norway}


\begin{abstract}
With the ever-increasing availability of data, there has been an explosion of interest in applying modern machine learning methods to fields such as modeling and control. However, despite the flexibility and surprising accuracy of such black-box models, it remains difficult to trust them. Recent efforts to combine the two approaches aim to develop flexible models that nonetheless generalize well; a paradigm we call Hybrid Analysis and modeling (HAM). In this work we investigate the Corrective Source Term Approach (CoSTA), which uses a data-driven model to correct a misspecified physics-based model. This enables us to develop models that make accurate predictions even when the underlying physics of the problem is not well understood. We apply CoSTA to model the Hall-H\'{e}roult process in an aluminum electrolysis cell. We demonstrate that the method improves both accuracy and predictive stability, yielding an overall more trustworthy model.
\end{abstract}

\end{frontmatter}


\section{Introduction/ motivation}
\label{sec:introduction}     

Many real-world phenomena can be modeled as differential equations, which allow us to predict the change in the state of the system over time. These equations are often derived from first principles, and we refer to the resulting models as \ac{PBM}. Through careful observation of physical phenomena, we can develop theories to describe and understand the underlying system. This understanding is condensed into mathematical equations, which can be solved to make predictions about the system. 
\ac{PBM}s have many inherent advantages. Due to their sound foundations from first principles, they are intuitive and explainable, they typically generalize well to situations where the assumptions are upheld, and there are mature theories that allow us to analyze their properties (e.g. stability and robustness to uncertainties and noise). 
However, accurately modeling many real-world systems comes at a high computational price. We may be forced to make assumptions to reduce the complexity of our model and minimize computational requirements. This is often necessary when developing control systems. We may also simply fail to accurately describe aspects of the observations. This can result in an incomplete, unfaithful, or overly simplified representation of the original system. 


\ac{DDM} is an alternative approach that does not base itself on an understanding of the physics; but instead attempts to approximate the underlying function directly from measurement data.
Over the past decade, the rapid progress in machine learning has created a massive demand for data, with a supply to match. 
This has enabled the development of \ac{DDM}s for a wide range of tasks, such as the identification of an aluminum electrolysis process \citep{MEGHLAOUI19981419}, and automatic contamination detection for polymer pellet quality control \citep{PENG2022107836}.
\ac{DDM}s offer enormous flexibility, and can often achieve remarkable accuracy with relatively little computation.
They can therefore be used even when we lack a complete understanding of the underlying physics of a system. 
However, it is well known that these models do not generalize well, meaning that they often fail when faced with data that is not well represented by the training data. 
Many classes of \ac{DDM}s also require unreasonably large amounts of data to be useful.
These drawbacks mean that when \ac{DDM}s are used in practice there is a preference for more transparent multivariate statistical models that can yield more insight into industrial processes, e.g. estimating the internal state of an aluminum smelting process during operation \citep{ABDMAJID20112457}.

\begin{figure}[ht]
    \centering
	\includegraphics[width=0.7\linewidth]{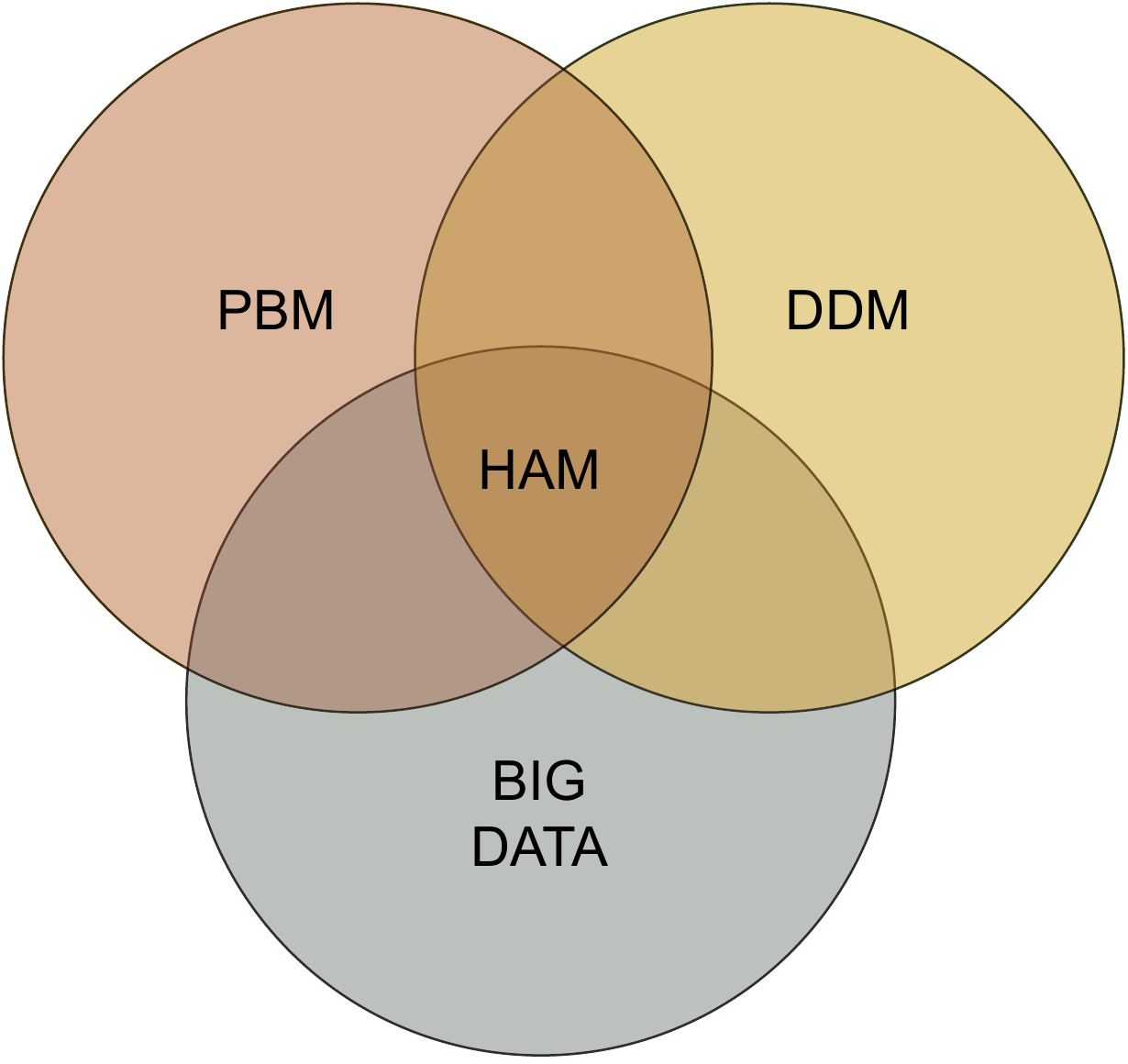}
	\caption{Hybrid analysis and modeling: working at the intersection of PBM, DDM and Big data.
	}
	\label{fig:ham} 
\end{figure}

In this work, we argue that combining the two approaches can help mitigate the disadvantages of both. 
Models following this paradigm are developed at the intersection of \ac{PBM}, \ac{DDM} and Big-Data (see Figure~\ref{fig:ham}). 
We call this approach \ac{HAM}, although many other terms have been coined in the literature, such as Informed Machine Learning \citep{ruden2019informed}, \ac{SciML} \citep{rackauckas_differentialequations.jl_2017}, and Structured Learning \citep{pineda_theseus_2022}. 
Here we list some of the main approaches to \ac{HAM}. 

The most straightforward class of methods is to simply embed a \ac{PBM} into a differentiable framework such as PyTorch \citep{paszke_pytorch_2019}. 
For example, recent work developed a differentiable convex optimization solver that can be used as a module in a \ac{NN}\citep{amos_2017_optnet}. 
The same has been done for linear complementarity problems, which has been used to create a differentiable physics simulator with analytical gradients \citep{avilabelbuteperes_2018_end}. 
This approach can serve as a powerful inductive bias for machine learning problems, allowing the specification of structure and constraints. 
However, \ac{PBM} methods are often iterative in nature, which increases computational costs relative to a standard \ac{NN} during both training and inference.
Instead of encoding prior knowledge to produce increasingly complex models, other approaches introduce inductive biases into the training method itself. 
The \ac{PINN} approach treats a \ac{NN} as the solution $\state$ of a \ac{PDE} \citep{raissi2019physics}, e.g. $\differentiate\state = \source(\state)$, where $\differentiate$ is a linear differential operator such that $\differentiate\state$ represents any linear combination of derivatives of $\state$. 
Using automatic differentiation, every term of $\differentiate\state$ can be computed for a selection of sample points, and the network will converge to the true solution $\state$ when optimized with the cost function $(\differentiate\state - \source(\state))^2$. 
This penalty term can be introduced as a soft constraint for models that are additionally trained on measurement data. 
In practice, optimizing such complex cost functions turns out to be quite challenging \citep{krishnapriyan_characterizing_2021}. Data-driven equation discovery (eg. \cite{vaddireddy2020fes}, \cite{RAVIPRAKASH2022107862}) is another approach that searches for equations that fit the data well. 
This is useful when the learned model needs to be interpretable and human-readable. This approach is arguably useful for scientific discovery as well, as it produces equations that can be further analyzed and combined with existing theory. 
A notable work is SinDy \citep{brunton_discovering_2016a}, which uses compressed sensing to approximate data using a sparse library of functions.
This has inspired other methods that search for sparse solutions (\cite{BAKARJI2021110219, Champion22445}), and other works optimize this search by trying to discover symmetries in the data \citep{udrescu2020ai}. 
These approaches have only been shown to work for relatively low dimensional examples, and require significant computational time. 
Deep symbolic regression approaches (\cite{kim_integration_2021} and \cite{XU2021110592}) treat a \ac{NN} itself as an expression tree and optimize it directly to obtain a closed form equation, where the neurons in each layer have different activation functions representing the library of allowed functions. 
While this can quickly fit higher dimensional data, like standard \ac{NN}s it tends to overfit the data. 
While deep symbolic regression can in theory express arbitrary compositions of the allowed functions, not all of these functions are relevant, and their presence can induce overfitting. 
A related concept, called \ac{PGNN} (\cite{pawar2021pgml, pawar2021msw,haakon2022pgnn}), mitigates this somewhat by only using features that appear in existing \ac{PBM}s, along with standard activation functions such as \texttt{ReLU} to retain the universal approximation capabilities of the network. 
These functions act as a store of prior knowledge that the network can utilize, while still modeling the unknown physics as a black box. 
Complicated features can increase the computational cost, similarly to the \ac{PBM} embedding approach discussed above.

While many \ac{HAM} approaches have seen some success, they suffer from a variety of issues such as increased computational cost for training and inference, difficult training convergence, and overfitting. See (\cite{9064519,10.1007/978-3-030-44584-3_43,ARIASCHAO2022107961,BRADLEY2022107898}) for more in-depth reviews of this field. In this work we investigate the efficacy of the \ac{CoSTA} approach, where the output of a discretized \ac{PBM} is corrected by a \ac{DDM} trained on the error of the base model. 
This approach is a natural way to use existing models. 
For example, \cite{lundbyanh2021} used compressed sensing to recover the residual of a PBM from sparse measurements, which is used to improve state estimates using a Kalman filter. \ac{CoSTA} also has theoretical justifications, as it possible to correct for a variety of model errors in this way, as shown by \cite{blakseth2022dnn}. 
\cite{blakseth2022dnn} demonstrated that CoSTA works for simple, one-dimensional heat transfer problems. In \citep{blaksethcpb2022}, the same work was extended to 2D and was also demonstrated that the CoSTA model has inbuilt sanity check mechanism. 

In this work we extend and apply \ac{CoSTA} to correct a misspecified \ac{PBM} of a complex aluminum extraction process simulation. The main contributions of this work that differentiates it from previous works are:
\begin{itemize}
    \item An extension of CoSTA to multidimensional problems: The previous works utilizing CoSTA were limited to modeling a single state temperature in either one or two-dimensional heat transfer. The current application of the aluminum extraction process involves eight states. 
    \item A successful application of CoSTA to a system with external control inputs: None of the previous work involved any control inputs. In the current work, five inputs are used to excite the system.  
    \item A successful application of CoSTA to a system with complex coupling between different states and inputs: The complex system considered here involves eight states and five inputs which form a set of eight ordinary equations which are highly coupled. The previous works involving heat transfer involved only one partial differential equations hence the potential of CoSTA to couple problems was never evaluated earlier.
\end{itemize}

This paper is structured as follows. Section~\ref{sec:theory} presents the relevant theory behind the aluminum extraction process, \ac{NN}s, and \ac{CoSTA}. We then outline the methodology of the work in Section~\ref{sec:Method_experiment}, namely how the data was generated, how the models were trained, and how they were evaluated. In Section~\ref{sec:resultsanddiscussion} we present present the results, and give a detailed discussion about the behaviour of the process and the models. We then summarise our findings and outline future work in Section~\ref{sec:conclusions}.

\section{Theory}
\label{sec:theory}

In order to investigate the applicability of \ac{CoSTA} to engineering applications we perform a case study on an aluminum extraction process using the Hall–H\'{e}roult process.
In the following sections we describe the underlying \ac{PBM} for this system, the fundamentals of \ac{NN}s, and the \ac{CoSTA} approach to \ac{HAM}.

\subsection{Physics-based model for aluminum extraction}
\label{subsec:pbm}
\begin{figure}[ht]
    \centering
    \includegraphics[width=\linewidth]{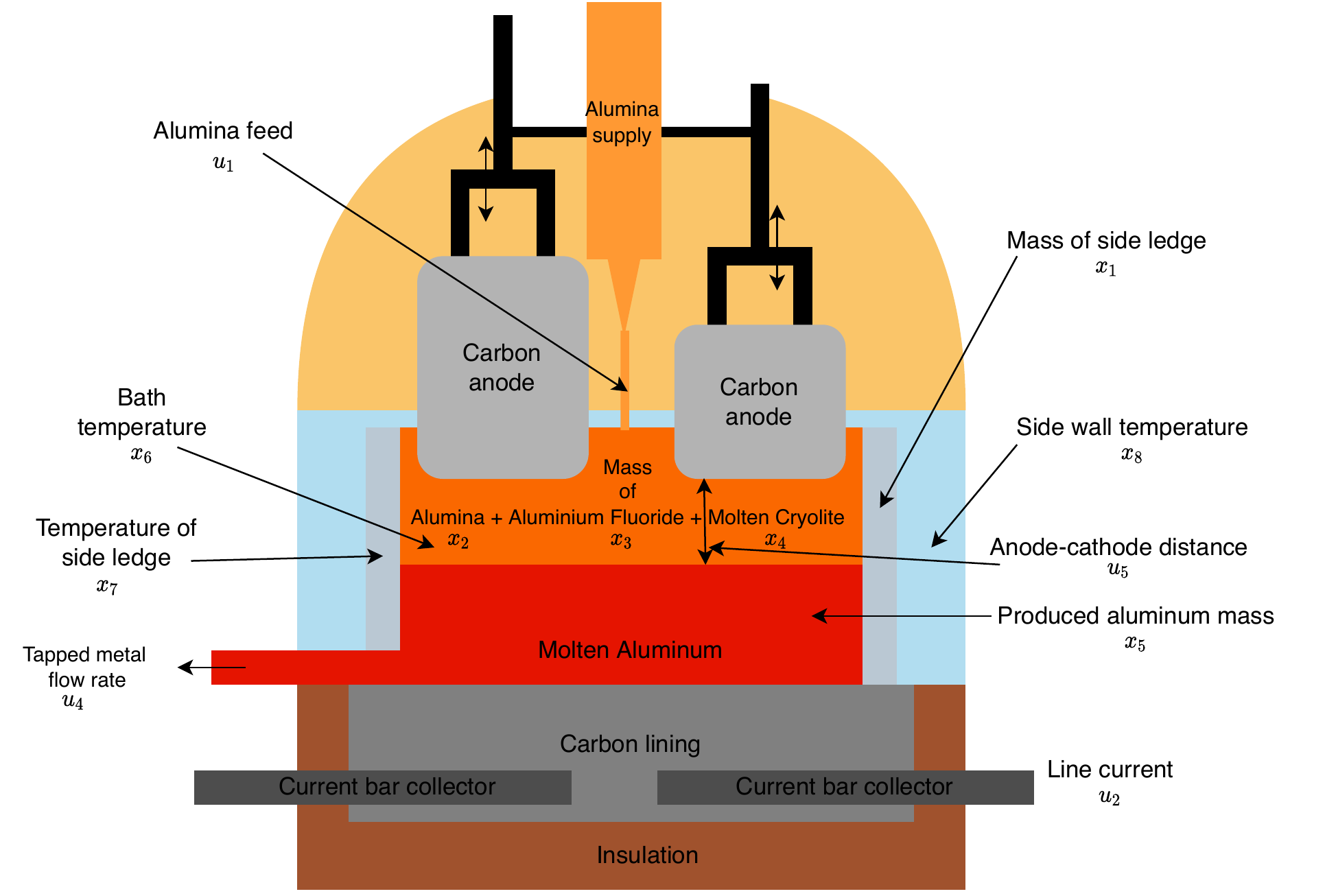}
    \caption{Schematic of the setup}
    \label{fig:schematicofthesetup}
\end{figure}
An overview of the physical plant is shown in Figure~\ref{fig:schematicofthesetup}. 
A \ac{PBM} of the plant can be derived from the mass/energy balance of the system.
We omit this step and present the model directly.
The internal dynamics of the aluminum electrolysis cell are described by a set of \ac{ODE}, with the general form:
\begin{equation}
    \Dot{\state} = \source(\state, \xinput),
    \label{eq:nonlin_state_space}
\end{equation}
where $\state \in \mathbb{R}^8$ is the state vector, $\xinput \in \mathbb{R}^5$ are external inputs, and $f(\state, \xinput)$ describes the nonlinear dynamics. 
Table~\ref{table:states_inputs} shows the names of the internal states and external inputs.
\begin{table}[ht]
    \begin{center}
    \caption{Table of states and inputs}
    \small
    \begin{tabular}{l|l|l}
    \hline
    Variable & Physical meaning & Unit  \\ \hline
    $x_1$ & mass side ledge & $kg$ \\
    $x_2$ & mass \ce{Al2O3} & $kg$  \\
    $x_3$& mass \ce{AlF_3} & $kg$ \\
    $x_4$ & mass \ce{Na_3 AlF_6} & $kg$  \\
    $x_5$ & mass metal & $kg$ \\
    $x_6$ & temperature bath & $^\circ C$ \\
    $x_7$ & temperature side ledge & $^\circ C$ \\
    $x_8$ & temperature wall & $^\circ C$\\
    \hline
    $u_1$ &  \ce{Al2O3} feed &$kg/s$ \\
    $u_2$ & Line current  & $kA$\\
    $u_3$ & \ce{AlF_3} feed & $kg/s$\\
    $u_4$ & Metal tapping & $kg/s$\\
    $u_5$ & Anode-cathode distance & $cm$ \\
    \hline
    \end{tabular}
    \label{table:states_inputs}
    \end{center}
\end{table}
The intrinsic properties of the \ce{Al2O3}+\ce{AlF_3}+\ce{Na_3 AlF_6} mixture are determined by the mass ratios of $x_2$ (\ce{Al2O3}) and $x_3$ (\ce{AlF_3}), written as:
\begin{equation}\label{eq:alu_ratios}
\begin{aligned}
    c_{x_2} &= x_2/(x_2 + x_3 + x_4)  \\
    c_{x_3} &= x_3/(x_2 + x_3 + x_4)
\end{aligned}
\end{equation}
We then define the following quantities:
\begin{subequations}\label{eq:alu_nonlin_help_fun}
\begin{align}
    g_1 &=  991.2 + 112 c_{x_3} + 61 c_{x_3}^{1.5} - 3265.5 c_{x_3}^{2.2} \label{eq:liquidus_temp}  \\
        &- \frac{793 c_{x_2}}{- 23 c_{x_2} c_{x_3} - 17 c_{x_3}^{2} + 9.36 c_{x_3} + 1} \nonumber \\
    g_2 &=  \text{exp}\,\left(2.496 - \frac{2068.4}{273+x_6} - 2.07c_{x_2}\right)\\
    g_3 &= 0.531 + 3.06 \cdot 10^{-18} u_{1}^{3} - 2.51 \cdot 10^{-12} u_{1}^{2} \\ &+ 6.96 \cdot 10^{-7} u_{1} 
        - \frac{14.37 (c_{x_2} - c_{x2,crit}) - 0.431}{735.3 (c_{x_2} - c_{x2,crit}) + 1}  \nonumber\\
    g_4 &= \frac{0.5517 + 3.8168 \cdot 10^{-6}u_2}{1 + 8.271 \cdot 10^{-6}u_2}\\
    g_5 &= \frac{3.8168\cdot 10^{-6} g_3 g_4 u_2}{g_2(1 -g_3)} 
\end{align}       
\end{subequations}
where $g_1$ is the liquidus temperature $T_{liq}$, $g_2$ is the electrical conductivity $\kappa$, $g_3$ is the bubble coverage , $g_4$ is the bubble thickness $d_{bub}$  and $g_5$ is the bubble voltage drop $U_{bub}$.
The critical mass ratio $c_{x_2, crit}$ is given in Table~\ref{table:const_alu_simulator}.

The full \ac{PBM} can now be written as a set of 8 \ac{ODE}s:
\begin{subequations}\label{eq:alu_equations}
\begin{align}
        \Dot{x}_1 &=\,\frac{k_1(g_1 - x_7)}{x_1 k_0} - k_2 (x_6 - g_1) \\
        \Dot{x}_2 &=\, u_1 - k_3 u_2\\
        \Dot{x}_3 &=\, u_3 - k_4 u_1\\
        \Dot{x}_4 &=\, -\frac{k_1 (g_1 - x_7)}{x_1 k_0} + k_2 (x_6 - g_1) + k_5 u_1\\
        \Dot{x}_5 &=\, k_6 u_2 - u_4 \\
        \Dot{x}_6 &=\, \frac{\alpha}{x_2+x_3+x_4} \Bigg[ u_2 g_5 + \frac{u_2^2 u_5}{2620 g_2} - k_7 (x_6 - g_1)^2 \\
                  & \, \;  + k_8 \frac{(x_6 - g_1)(g_1 - x_7)}{k_0x_1} - k_9 \frac{x_6 - x_7}{k_{10} + k_{11} k_0  x_1} \Bigg] \nonumber\\
        \Dot{x}_7 &=\, \frac{\beta}{x_1} \biggl[ \frac{k_{9} (g_1 - x_7)}{k_{15}k_0 x_1} - k_{12}(x_6 - g_1)(g_1 - x_7)  \\
                  & \, \; +  \frac{k_{13}(g_1 - x_7)^2}{k_0x_1} 
                  \,  - \frac{x_7 - x_8}{k_{14} + k_{15}k_0  x_1} \biggr]\nonumber\\
        \Dot{x}_8 &=\, k_{17} k_9 \left(\frac{x_7 - x_8}{k_{14} + k_{15} k_0 \cdot x_1} - \frac{x_8 - k_{16}}{k_{14} + k_{18}}\right)
\end{align}
\end{subequations}
The constants $(k_0,\; ..,\; k_{18}, \; \alpha, \; \beta)$ in Equation~\eqref{eq:alu_equations} are described and given numerical values in Table~\ref{table:const_alu_simulator}. 

\begin{figure}[ht]
	\includegraphics[width=\linewidth]{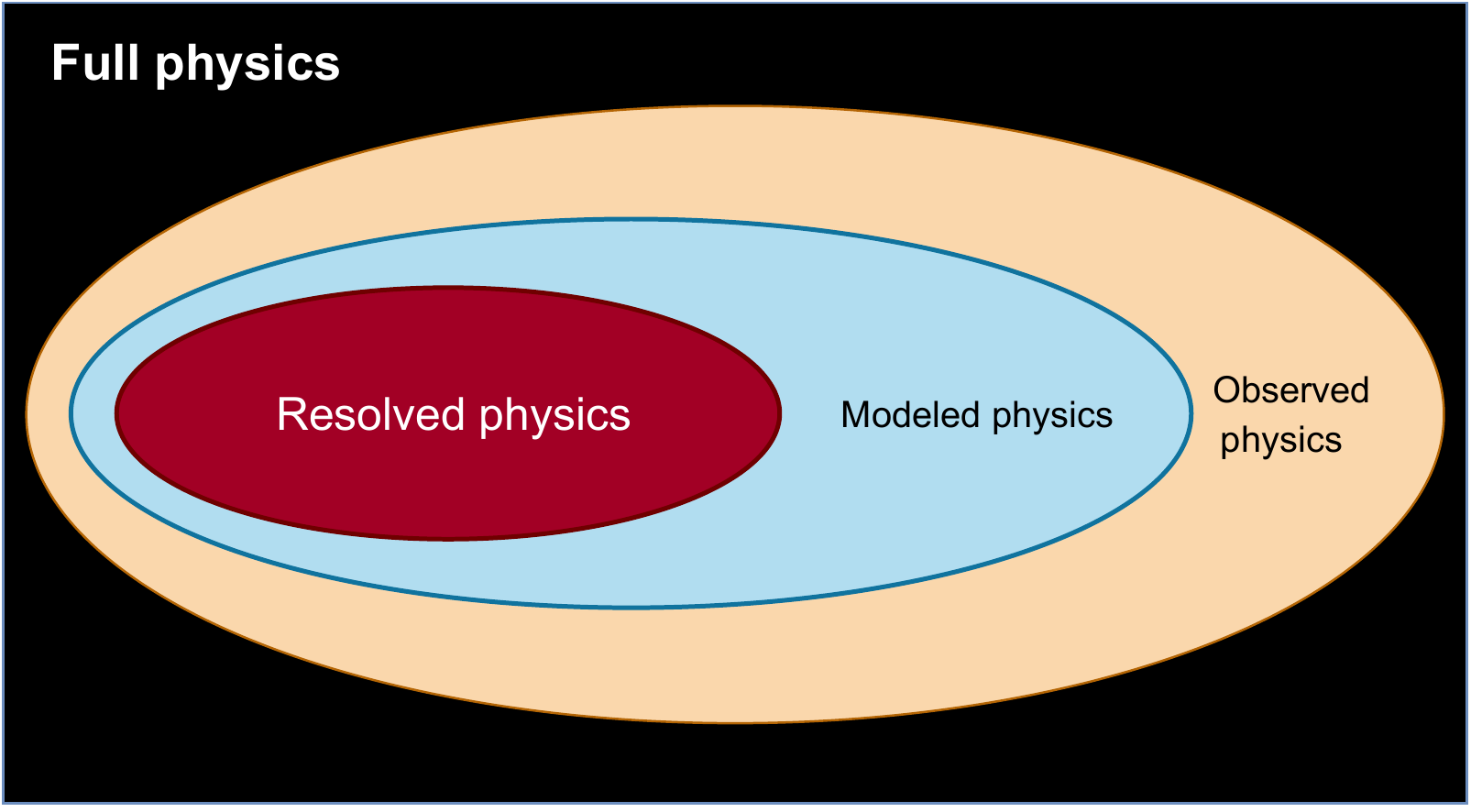}
        \caption{PBM: black part corresponds to unknown / unmodeled physics, orange ellipse corresponds to observed physics, purple ellipse corresponds to actually modelled physics while red ellipse signifies the actual physics solved for.}
	\label{fig:pbm} 
\end{figure}

As previously discussed, we are interested in modeling scenarios where the \ac{PBM} does not capture the full underlying physics of the system. This is illustrated in Figure~\ref{fig:pbm}, where the black background represents physics that cannot be observed or are not adequately explained by available theory. The orange ellipse represent physics ignored due to assumptions. The red ellipse corresponds to resolved physics after solving \ac{PBM} numerically, while the blue ellipse correspond to the modeled physics.

The model presented in Equation~\ref{eq:alu_equations} makes some simplifications compared to the real process of aluminum electrolysis.
Firstly, we only model the heat transfer through the side walls, with the assumption that heat flow through the top and bottom of the plant is negligible in comparison.
The model may thus overestimate the internal temperatures, and the required power input through the line current $u_2$ may be slightly lower than in practice. 
Secondly, the spatial variations of the state variables are not considered. 
Instead, only the average values of the states such as the side ledge temperature, or cumulative values such as the mass of side ledge $x_1$ are computed. 
Routine operations such as the alumina feeding and anode replacement disturbs local thermal balance and causes local thermal imbalances \citep{cheung2013spatial}.
Modeling these local variations would require knowledge of the mass transfer inside the cell due to the flow patterns and velocity fields in the bath, current distribution etc. 
These phenomena (corresponding to the orange ellipse of Figure~\ref{fig:pbm}) are very difficult to model and measure, and are therefore omitted to reduce complexity. 
A more detailed derivation the model can be found in \citep{lundby_2ndpaper}.

For this case study, we choose to use simulation data generated from Equation~\eqref{eq:alu_equations} in order to validate the \ac{CoSTA} method. To that end, we make a further simplification: we ignore Equation~\eqref{eq:liquidus_temp} and set the liquidus temperature $g_1$ to a constant. 
\newcommand{\gconst}[0]{{g_{1, PBM}}}
\begin{equation}
    g_{1, PBM} = 968 ^{\circ}C.
\end{equation}
We refer to the resulting model as the \textit{ablated \ac{PBM}}. 
This choice was made because the model is particularly sensitive to errors in $g_1$.
Inspecting Equation~\eqref{eq:alu_equations} shows that the ablated \ac{PBM} will incorrectly predict the evolution of $[x_1, x_4, x_6, x_7, x_8]$. 
As we will see later in Section~\ref{sec:resultsanddiscussion} and Figure~\ref{fig:Rolling_forecast}, this can lead to errors of roughly $\si{5\degreeCelsius}$ in $g_1$, and $\si{500\kilogram}$ in the side ledge mass $x_1$ (a relative error of $10\%$).
The aim of the case study is to develop a \ac{DDM} to correct the ablated \ac{PBM} using measurement data sampled from the true model. 

%
\begin{table}[ht]
    \begin{center}
    \caption{Constants in the simulator}
    \small
    \begin{tabular}{l|l|l}
        \hline
        Constant &Physical meaning &Numeric value  \\ \hline
        $k_0$ & $1/(\rho_{sl}A_{sl})$ &$2\cdot 10^{-5}$ \\
        $k_1$ &$2k_{sl}A_{sl}/\Delta_{fus} H_{cry}$ &$7.5\cdot 10^{-4}$ \\
        $k_2$ &$h_{bath-sl}A_{sl}/\Delta_{fus} H_{cry}$ &$0.18$  \\
        $k_3$& $0.002\frac{M_{Al_2O_3}\cdot CE}{z\cdot F}$ &$1.7\cdot 10^{-7}$ \\
        $k_4$ & $C_{Na_2O}\frac{4M_{AlF_3}}{3M_{Na_2O}}$ &$0.036$  \\
        $k_5$ & $C_{Na_2O}\frac{2M_{cry}}{3M_{Na_2O}}$ &$0.03$  \\
        $k_6$ & $0.002\frac{M_{Al}\cdot CE}{z\cdot F}$&$4.43\cdot 10^{-8}$ \\
        $k_7$ & $k_2 \cdot c_{p_{cry, \; liq}}$ &$338$ \\
        $k_8$ & $k_1 \cdot c_{p_{cry, \; liq}}$ & $1.41$ \\
        $k_9$ & $A_{sl}$ &$17.92$ \\
        $k_{10}$ & $1/h_{bath-sl}$ &$0.00083$ \\
        $k_{11}$ & $1/(2k_{sl})$ &$0.2$ \\
        $k_{12}$ & $k_2\cdot c_{p_{cry,\; s}}$ &$237.5$ \\
        $k_{13}$ & $k_1\cdot c_{p_{cry,\; s}}$&$0.99$ \\
        $k_{14}$ & $x_{wall}/(2k_{wall})$ &$0.0077$ \\
        $k_{15}$ & $1/(2k_{sl})$ &$0.2$ \\
        $k_{16}$ & $T_{0}$ &$35$ \\
        $k_{17}$ & $1/(m_{wall}c_{p, \; wall})$ &$5.8 \cdot 10^{-7}$ \\
        $k_{18}$ & $1/h_{wall-0}$ &$0.04$ \\
        $\alpha$ & $1/c_{p_{bath, \; liq}}$ &$5.66\cdot 10^{-4}$\\
        $\beta$ & $1/c_{p_{cry, \; sol}}$ &$7.58\cdot 10^{-4}$\\
        $c_{x_{2, crit}}$ & \todo{Equation here} & $0.022$\\
        \hline
    \end{tabular}
    \label{table:const_alu_simulator}
    \end{center}
\end{table}

\subsection{Data-driven modeling using neural networks}
\label{subsec:ddm}
\begin{figure}[ht]
	\includegraphics[ width=\linewidth]{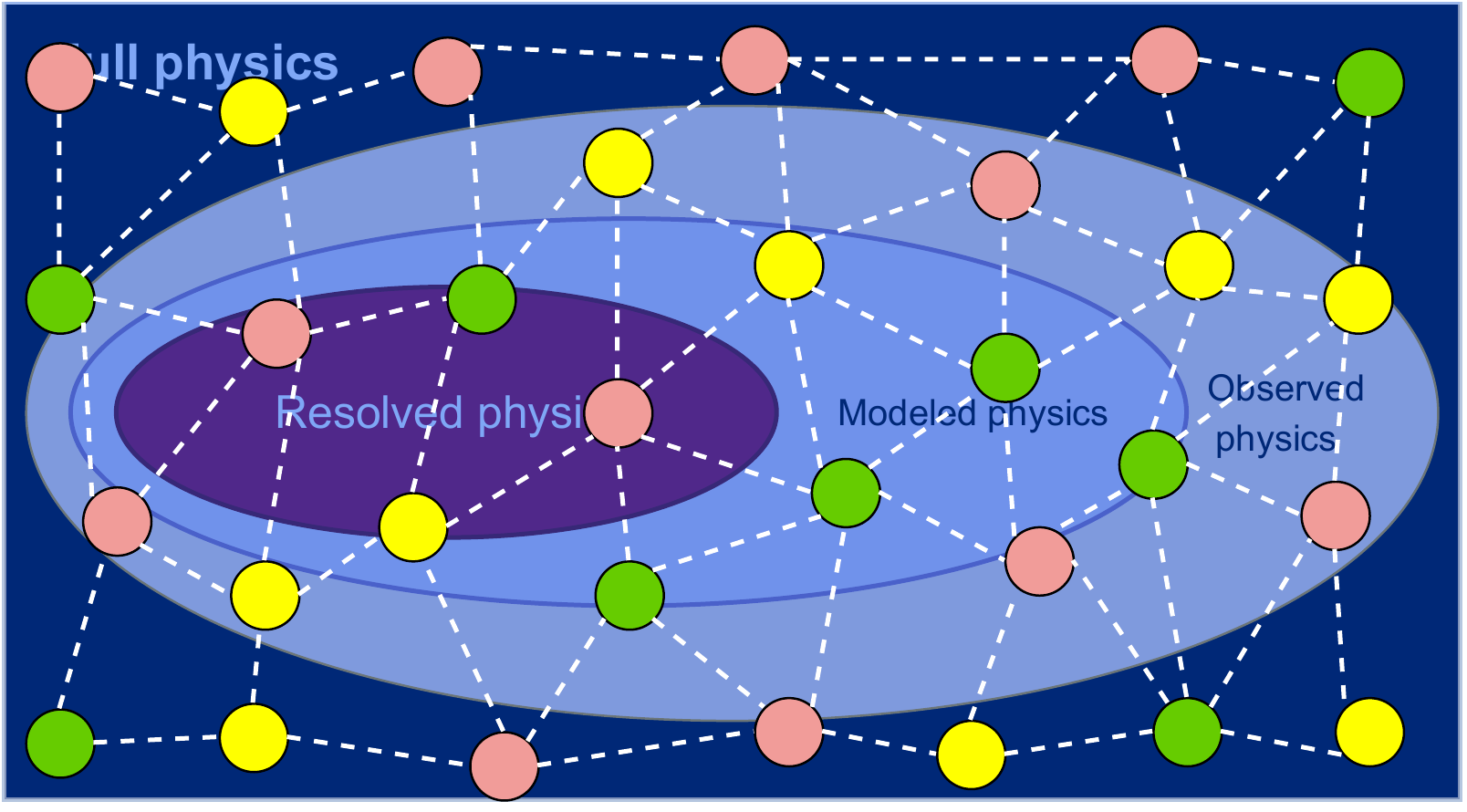}
        \caption{\ac{DDM}: It is assumed that the data is a realization of the true data-generating process, models trained on the         data will implicitly capture the full physics of this process. 
	}
	\label{fig:ddm} 
\end{figure}
Instead of making assumptions and fitting our theories to the data, \ac{DDM}s can learn to approximate the underlying process directly from data. Figure~\ref{fig:ddm} shows this conceptually. In this work we focus on modeling Equation~\eqref{eq:alu_equations} using a \ac{NN}. A \ac{NN} can be seen as a general function approximator. We denote the trainable parameters of the model as $\boldsymbol{\theta} \in \mathbb{R}^p$ and denote the network as
\begin{equation}
    \netoutput = \net(\netinput; \netparameters),
\end{equation}
where $\netinput \in \mathbb{R}^d, \netoutput \in \mathbb{R}^s$ are the inputs and outputs to the model respectively. The network is composed of several layers. The $j$th layer operates on the output of the previous layer, and produces its own output which we call $\mathbf{Z}^{j} \in \mathbb{R}^{L_j}$. A \textit{fully connected layer} can be seen as an affine transformation composed with a nonlinear activation function $\sigma: \mathbb{R}^n \mapsto \mathbb{R}^n$
\begin{equation}
    \mathbf{Z}^{j} = \sigma (\netweights{j} \mathbf{Z}^{j-1} + \netbiases{j}),
\end{equation}
where $\netweights{j} \in \mathbb{R}^{L_{j} \times L_{j-1}}$ is called the \textit{weight or connection matrix}, and $\netbiases{j} \in \mathbb{R}^{L_{j}}$ is the \textit{bias vector} of layer $j$. We denote the rows of $\netweights{j}$ as $\netweight{j}{i}$, and the individual bias terms as $\netbias{j}{i}$. The nonlinear activation function $\sigma$ can, for example, be the \texttt{sigmoid} function, hyperbolic tangent function (\texttt{tanh}) or the binary step function to mention a few. All of these operate element-wise over $\mathbf{Z}^j$, but there exist functions that operate on groups on elements, for example the \texttt{maxout} activation function. The most popular activation function in recent times is \texttt{ReLU}, due to its computational simplicity, representational sparsity and non-vanishing gradients. The \texttt{ReLU} activation function is given by:
\begin{equation}
    \sigma(z) = \text{max}\{0,\;z\}. 
\end{equation}

From the previous section it can be seen that \ac{NN}s are dense models with many parameters. In fact, the largest networks in use today often have more parameters than the amount of available data to train them on. For example, the widely publicized GPT-3 model has 175 billion parameters \citep{brown_language_2020}. Because of this, avoiding overfitting and getting deep learning models to generalize is an important topic in deep learning, and methods that accomplish this are generically referred to as \textit{regularization} \citep{goodfellow2016deep}. Examples of such methods include weight decay \citep{krogh1991simple}, dropout \citep{srivastava_dropout_2014}, and batch normalization \citep{ioffe_batch_2015}, all of which are important tools in ensuring a low generalization error for these models. In recent years, more and more research has shifted towards sparse architectures with significantly fewer non-zero trainable parameters than their dense counterparts \citep{hoefler2021sparsity}. There are many reasons for this. First of all, sparser networks are much cheaper to store and evaluate. This is vital for practitioners wishing to deploy their models on lower cost hardware \citep{sandler_mobilenetv2_2018}. Secondly, recent work shows a tantalising hint that sparse models may in fact generalise better than their dense counterparts. In their seminal work, \cite{frankle2018lottery} show with high probability that randomly initialized dense \ac{NN}s contain subnetworks that can improve generalization compared to the dense networks. 

Many regularization methods can be expressed as a penalty function $R(\weights)$ that operates on the parameters $\netparameters$ of the network. The total loss function $\totalloss(\netinput_i, \netoutput_i, \netparameters)$ used for training the network can then be written as
\begin{equation}\label{eq:total-loss}
    \totalloss(\netinput_i, \netoutput_i, \netparameters) =  \loss(\netoutput_i, \net(\netinput_i; \netparameters)) + \lambda \wpenalty(\weights),
\end{equation} 
where the set $ {\cal D} = \{(\netinput_i, \netoutput_i)\}_{i=1}^N$ is the training dataset, $\loss(\cdot, \cdot)$ is the \textit{loss function} and $\lambda \in \mathbb{R}^{+}$ serves to trade-off $\loss(\cdot, \cdot)$ and $\wpenalty(\cdot)$. 

The standard choice of loss function $\loss(\cdot, \cdot)$ for regression tasks is the \ac{MSE}:
\begin{equation}\label{eq:mse}
    \loss(\netinput_i, \netoutput_i) =  (\netinput_i - \netoutput_i)^2.
\end{equation} 
In the training process, the total cost function $\totalloss(\cdot, \cdot)$ is minimized to find optimal values of the parameters:
\begin{equation}
    \begin{aligned}
    \netparameters^* = \underset{\netparameters}{\mathrm{argmin}}\left\{\frac{1}{N}\sum_{i=1}^N \totalloss(\netinput_i, \netoutput_i, \netparameters)   \right\}.
    \end{aligned}
    \label{eq:general_opt_NN}
\end{equation}
The most straightforward way to penalise non-sparse $\boldsymbol{\theta}$ is the $\ell_0$ norm, often referred to as the sparsity norm:
\begin{equation}
    \wpenalty_{\ell_0}(\weights) = ||\weights||_0 = \sum_i 
    \begin{cases}
     1 & w_i \neq 0,\\
     0 & w_i = 0.
    \end{cases}
\end{equation}
It is clear that $\ell_0(\netparameters)$ simply returns the number of nonzero parameters. It has been shown that adding this regularization term can yield unique solutions for over-determined linear systems, which is the basis of compressed sensing \citep{boche_compressed_2015}. However, $\ell_0(\netparameters)$ is non-differentiable, making it unsuitable for gradient descent optimization. In fact, \cite{natarajan1995sparse} show that this optimization problem is NP-hard \citep{natarajan1995sparse}. Instead, we can utilise the $\ell_1$ norm, which is a convex relaxation of the $\ell_0$ norm and is given by:
\begin{equation}
    R_{\ell_1}(\weights) = ||\weights||_1 = \sum_i|w_i|.
\end{equation}
The $\ell_1$ norm sometimes does not reduce the weights to zero, but rather to very small magnitudes. In this case, we can apply a threshold to the weights, and set all weights below this threshold to zero. This method is known as \textit{magnitude pruning} and is the simplest of a family of pruning methods \citep{hoefler2021sparsity}. Despite it's simplicity, it can reduce the computation complexity of a \ac{NN} while maintaining the performance of the model \citep{gale2013rbf}.

\subsection{Corrective source term approach (CoSTA)}
\label{subsec:CoSTA}
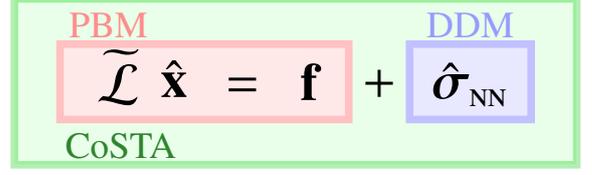
\begin{figure}
	\centering 
	\begin{tikzpicture}[scale=0.2]
	    
	    \draw[color=white, fill=green!15, opacity=0.8] (-18, -5.5) rectangle (19, 5.5);
	    \draw[line width=0.1cm, color=green!80!black, fill=green!0, opacity=0.3] (-18, -5.5) rectangle (19, 5.5);
	    
	    \draw[line width=0.1cm, color=white,fill=white] (-15, -2.5) rectangle (4, 2.5);
	    \draw[line width=0.1cm, color=red!40,fill=red!15,opacity=0.6] (-15, -2.5) rectangle (4, 2.5);
	    
	    \draw[line width=0.1cm, color=white, fill=white] (8, -2.5) rectangle (16, 2.5);
	    \draw[line width=0.1cm, color=blue!40,fill=blue!15,opacity=0.6] (8, -2.5) rectangle (16, 2.5);
	    
	    \node[scale=0.7*2, opacity=0.8] at (-11, -4.25)  {\textcolor{green!40!black}{CoSTA}};
	    \node[scale=0.7*2] at (-11.8, 3.75)  {\textcolor{red!45}{PBM}};
	    
	    \node[scale=0.7*2] at (12, 3.75)  {\textcolor{blue!40}{DDM}};
	    
	    \node[scale=2] at (-11, 0.3) {$\widetilde\differentiate$};
	    \node[scale=2] at (-7.5, 0.2) {$\hat{\state}$};
	    \node[scale=2] at (-3, -0.4) {$=$};
	    \node[scale=2] at (1.5, -0) {$\source$};
	    \node[scale=2] at (6, -0.1) {$+$};
	    \node[scale=2] at (12, -0.2) {$\hat{\residual}_{\textsc{nn}}$};
        
    \end{tikzpicture}
    \caption{CoSTA combines PBM and DDM into a unified model by adding a \ac{NN}-generated corrective source term to the governing equation of the PBM.}
    \label{fig:CoSTA_overview}
\end{figure}
In this section we outline the \ac{CoSTA} approach, illustrated in Figure~\ref{fig:CoSTA_overview}. Suppose we want to solve the following general problem:
\begin{alignat}{3}
    \label{eq:exact_in}
    \differentiate \state &= \source(\state,\xinput)
\end{alignat}
where $\differentiate$ is a differential operator, $\state$ is the unknown state of the system that we wish to compute, and $\source(\cdot, \cdot)$ is a source term that depends on the state $\state$ and external inputs $\xinput(t)$.

Assume now that we have a \ac{PBM} designed to predict $\state$, and let $\tilde{\state}$ denote the \ac{PBM}'s prediction of the true solution $\state$.
If $\tilde{\state}\neq \state$, there is some error in the \ac{PBM}, and this error must stem from at least one of the following misspecifications in the model:
\begin{enumerate}
    \item Incorrect $\source$ in Equation~\eqref{eq:exact_in}, replaced by $\tilde{\source}$.
    \item Incorrect $\differentiate$ in Equation~\eqref{eq:exact_in}, replaced by $\widetilde\differentiate$.
    \item A combination of the above.
    \item Discretization of $\differentiate$, replaced by $\differentiate_\mathcal{D}$\footnote{Derived using, for example, finite differences. This is necessary when Equation~\eqref{eq:exact_in} lacks analytical solutions, which is almost always the case.}. 
\end{enumerate}
Note that case 4 is also mathematically equivalent to misspecifying $\differentiate$. For example, $\frac{\partial}{\partial t}$ could be approximated using a forward finite difference. We can write this using the difference operator $\Delta_h$, such that $h$ is the time step and $\frac{1}{h}\Delta_h\,f(t) = (f(t+h)-f(t))/h$. We can therefore limit our discussion to Cases 1 and 2 without loss of generality.

Suppose now that the PBM-predicted solution $\tilde{\state}$ is given as the solution of the following system:
\begin{alignat}{3}
    \label{eq:PBM_in}
    \widetilde\differentiate \tilde{\state} &= \tilde{\source}
\end{alignat}
This formulation encompasses both Case 1 ($\widetilde\differentiate = \differentiate$ and $\tilde{\source} \neq \source$), Case 2 ( $\widetilde\differentiate \neq \differentiate$ and $\tilde{\source} = \source$), and combinations thereof (for $\widetilde\differentiate \neq \differentiate$ and $\tilde{\source} \neq \source$).
Furthermore, suppose we modify the system above by adding a source term $\hat{\sigma}$ to Equation~\eqref{eq:PBM_in}, and let the solution of the modified system be denoted $\hat{\tilde{\state}}$. Then, the modified system reads
\begin{alignat}{3}
    \label{eq:PBM_mod_in}
    \widetilde\differentiate \hat{\tilde{\state}} &= \tilde{\source} + \hat{\residual}
\end{alignat}
and the following theorem holds.

\paragraph{Theorem} Let $\hat{\tilde{\state}}$ be a solution of Equations~\eqref{eq:PBM_mod_in}, and let $\state$ be a solution of Equations~\eqref{eq:exact_in}. Then, for both operators $\widetilde\differentiate$, $\differentiate$ and both functions $\source$, $\tilde{\source}$, such that $\hat{\tilde{\state}}$ and $\state$ are uniquely defined, there exists a function $\residual$ such that $\hat{\tilde{\state}} = \state$.

\begin{figure}[ht]
	\includegraphics[width=\linewidth]{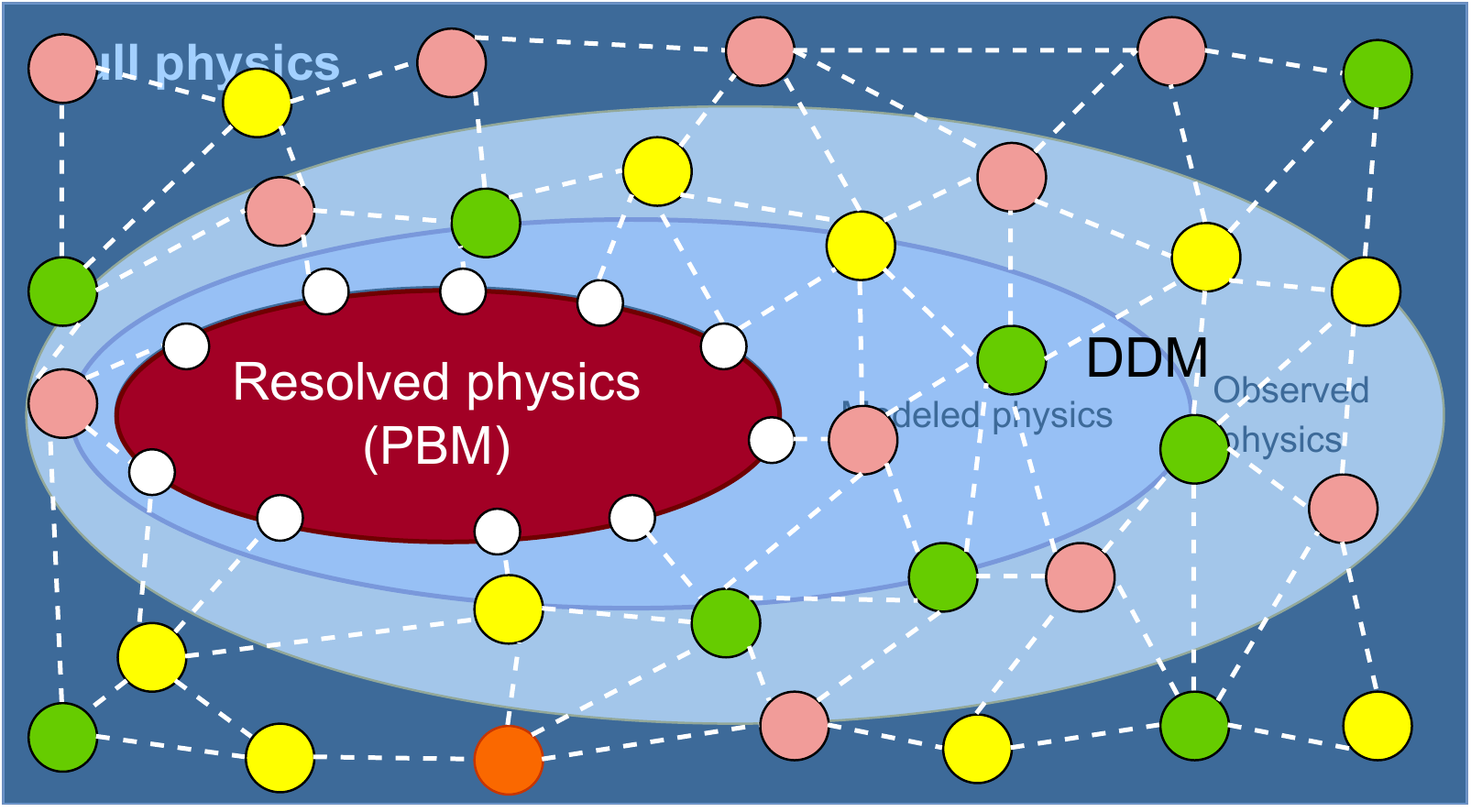}
	\caption{CoSTA: It maximizes the utilization of the well known PBM while correcting for the unknown using DDM. In the CoSTA, PBM is described by the set of differential equations describing the state of the system after the introduction of errors as explained in Section \ref{subsec:datageneration}. We call it resolved physics and represent it by the red ellipse. All the unaccounted physics and unintended numerical discretization errors are captured using the data-driven corrective source terms.
	}
	\label{fig:CoSTA} 
\end{figure}

\textit{Proof}:
Define the residual $\residual$ of the PBM's governing equation~\eqref{eq:PBM_in} as\footnote{Instead of defining the residual in terms of the approximate solution (e.g.\ as is done in truncation error analysis \citep[chapter~8]{leveque2002fvm}), we define $\residual$ by inserting the true solution into  Equation~\eqref{eq:exact_in}. Our proof is simpler, and fit well with real systems where state measurements are more readily available than the true governing equations.}
\begin{equation}
    \residual = \widetilde\differentiate \state - \tilde{\source}.
    \label{eq:residual}
\end{equation}
If we set $\hat{\residual} = \residual$ in Equation~\eqref{eq:PBM_mod_in}, we then obtain
\begin{align}
    \widetilde\differentiate \hat{\tilde{\state}} &= \tilde{\source} + \hat{\residual} \\
    &= \tilde{\source} + \widetilde{\differentiate} \state - \tilde{\source} \\
    &= \widetilde{\differentiate}\state 
    \\
    \implies \quad \hat{\tilde{\state}} &= \state + \mathbf{c}
\end{align}
where $\mathbf{c}$ is a function of independent variables. We can eliminate $\mathbf{c}$ by setting appropriate boundary conditions… 
\hfill $\blacksquare$\\
This shows that we can always find a corrective source term $\hat{\residual}$ that compensates for any error in the PBM's governing equation~\eqref{eq:PBM_in} such that the solution $\hat{\tilde{\state}}$ of the modified governing equation~\eqref{eq:PBM_mod_in} is equal to the true solution $\state$. 
This observation is the principal theoretical justification of CoSTA. 
As illustrated by Figure~\ref{fig:CoSTA}, the \ac{CoSTA} approach should be applicable to many physical problems that can be described using differential equations. 
%

\section{Method and experimental setup}
\label{sec:Method_experiment}
In this section we explain how we generated the data, how the data was divided into training, validation and test sets, and how the models were evaluated.

\subsection{Data generation and preprocessing}
\label{subsec:datageneration}
The dynamical system data is generated by integrating the set of non-linear \ac{ODE}'s in Eq.\eqref{eq:alu_equations} representing the system dynamics using the fourth-order numerical integrator \ac{RK4} with a fixed timestep $\Delta T = 10s$. One time-series simulation starts at an initial time $t_0$ with a set of initial conditions $\state(t_0)$, and last until a final time $T=5000\times \Delta T$. For the slow dynamics of the aluminum process, a sampling time of $10s$ turns out to be sufficiently fast with negligible integration errors. Higher sampling frequencies would lead to unnecessary high computational time and large amounts of simulation data. The initial conditions for each trajectory were uniformly sampled from the ranges shown in Table~\ref{table:init_conditions_aluminum}. Each simulation generates a set of trajectories consisting of $8$ states and $5$ inputs. 40 simulated trajectories are used for training the models, and 100 simulated trajectories are used as the test set. This relatively large number of test cases was to chosen to allow us to explore the statistics of how the model performs.
\begin{figure*}
     \centering
    \begin{subfigure}[t]{0.32\linewidth}
        \raisebox{-\height}{\includegraphics[width=\linewidth]{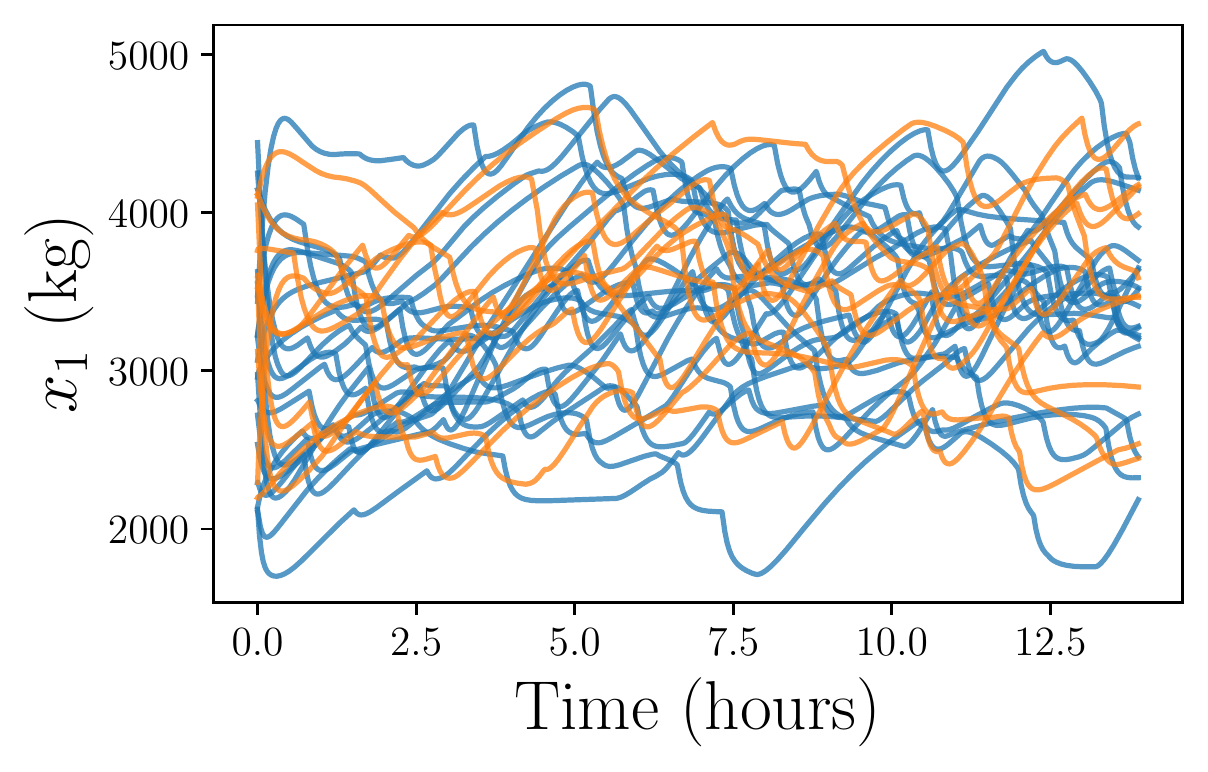}}
        \caption{Side ledge mass $x_1$}
        \label{subfig:training_testset_x1}
    \end{subfigure}
    \begin{subfigure}[t]{0.32\linewidth}
        \raisebox{-\height}{\includegraphics[width=\linewidth]{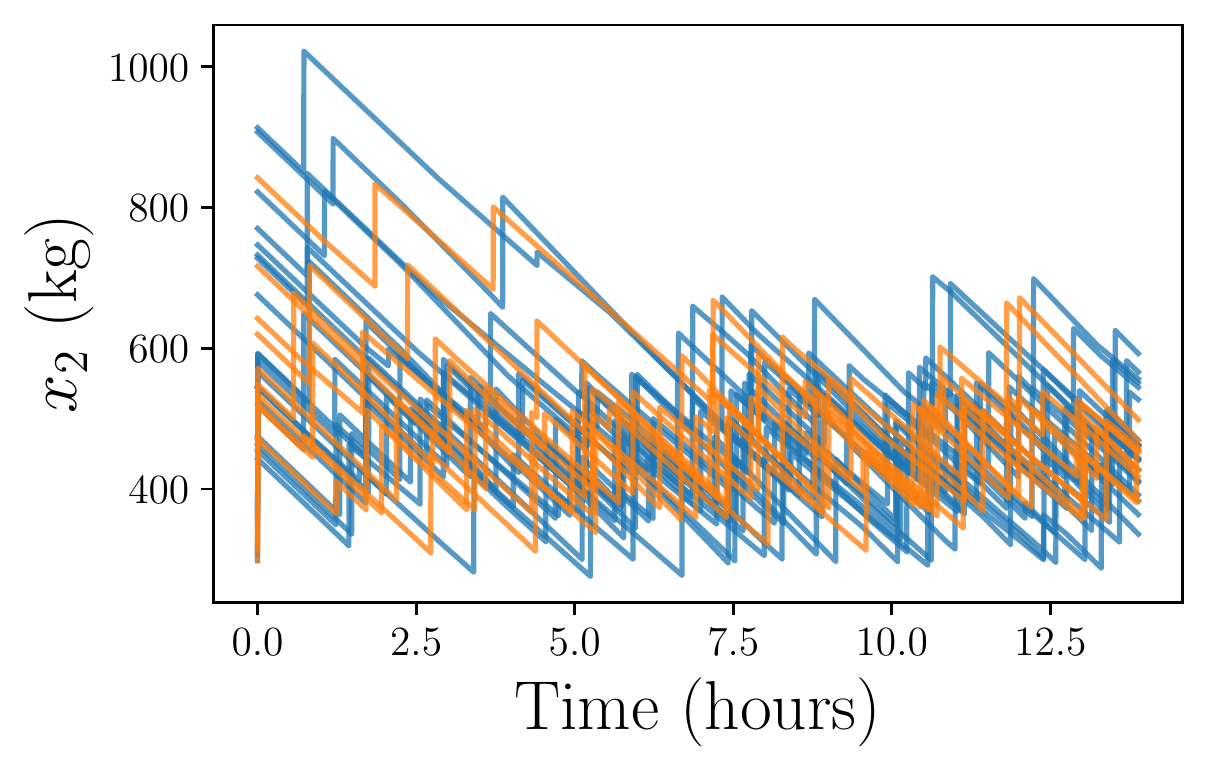}}
        \caption{Alumina mass $x_2$}
        \label{subfig:training_testset_x2}
    \end{subfigure}
    \begin{subfigure}[t]{0.32\linewidth}
        \raisebox{-\height}{\includegraphics[width=\linewidth]{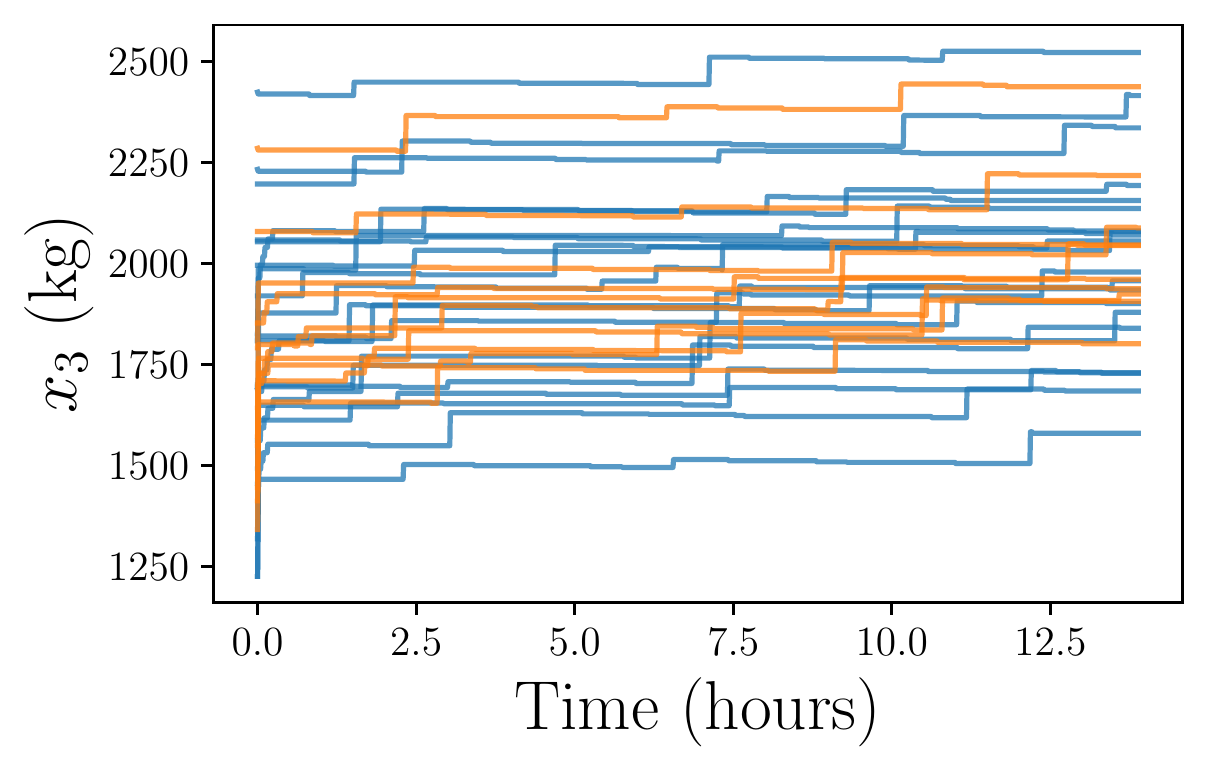}}
        \caption{Aluminum fluoride mass $x_3$}
        \label{subfig:training_testset_x3}
    \end{subfigure}
    \begin{subfigure}[t]{0.32\linewidth}
        \raisebox{-\height}{\includegraphics[width=\linewidth]{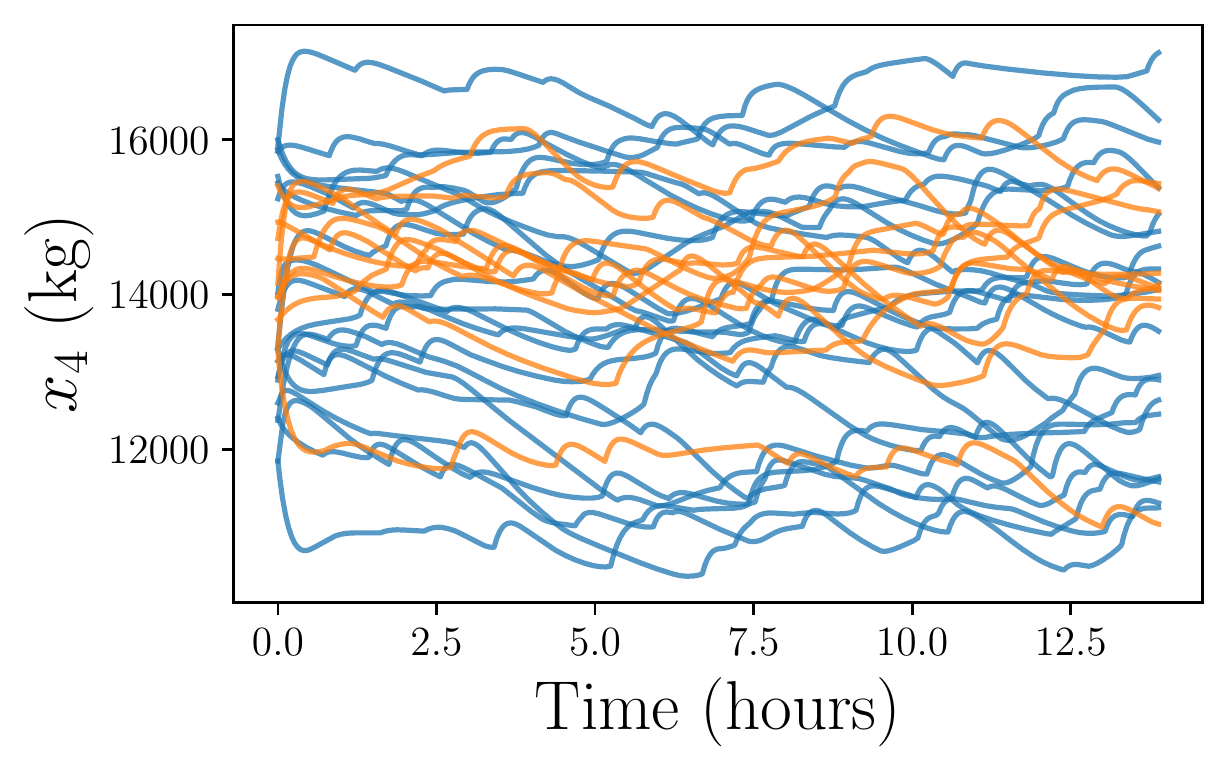}}
        \caption{Molten cryolite mass $x_4$}
        \label{subfig:training_testset_x4}
    \end{subfigure}
    \begin{subfigure}[t]{0.32\linewidth}
        \raisebox{-\height}{\includegraphics[width=\linewidth]{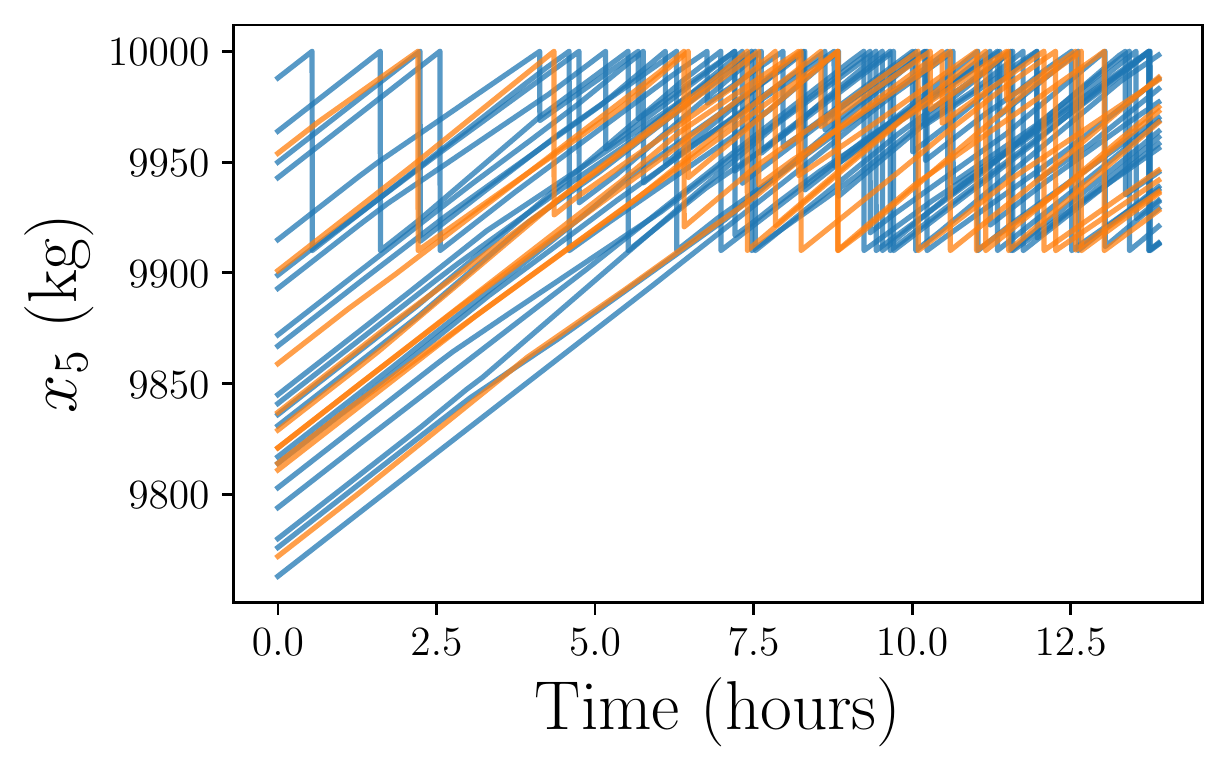}}
        \caption{Produced aluminum mass $x_5$}
        \label{subfig:training_testset_x5}
    \end{subfigure}
    \begin{subfigure}[t]{0.32\linewidth}
        \raisebox{-\height}{\includegraphics[width=\linewidth]{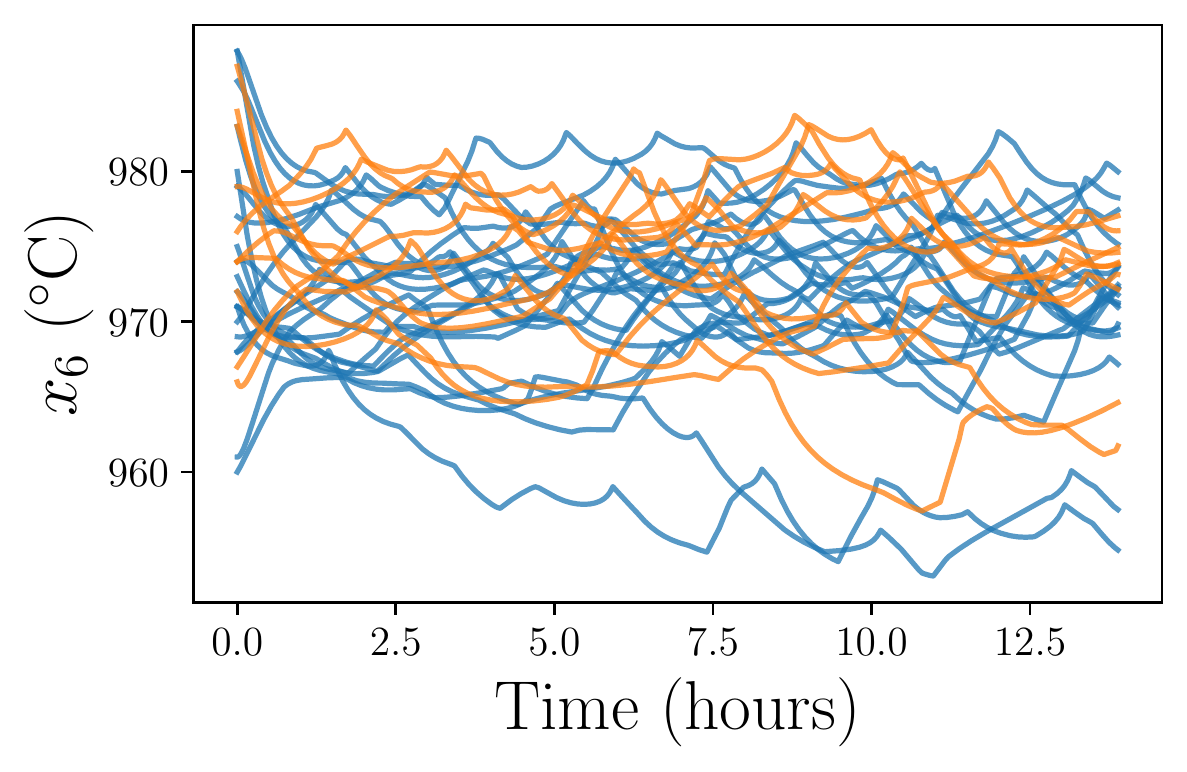}}
        \caption{Bath temperature $x_6$}
        \label{subfig:training_testset_x6}
    \end{subfigure}
    \begin{subfigure}[t]{0.32\linewidth}
        \raisebox{-\height}{\includegraphics[width=\linewidth]{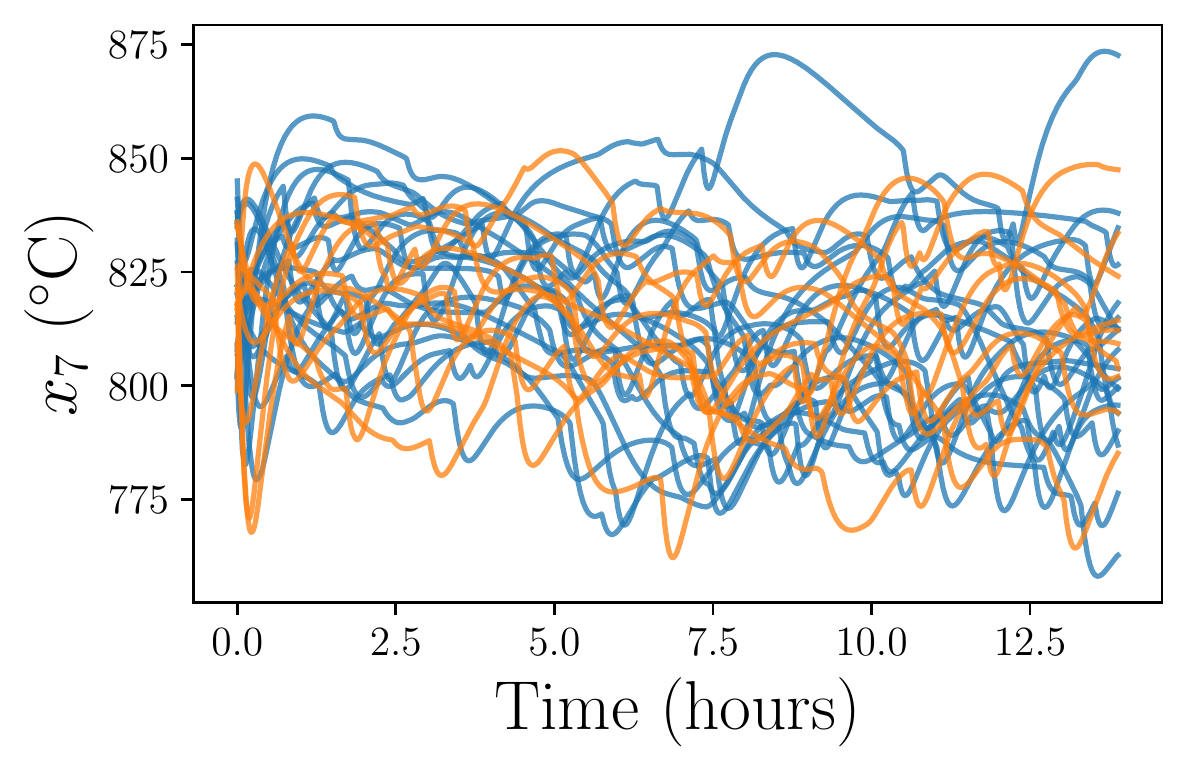}}
        \caption{Side ledge temperature $x_7$}
        \label{subfig:training_testset_x7}
    \end{subfigure}
    \begin{subfigure}[t]{0.32\linewidth}
        \raisebox{-\height}{\includegraphics[width=\linewidth]{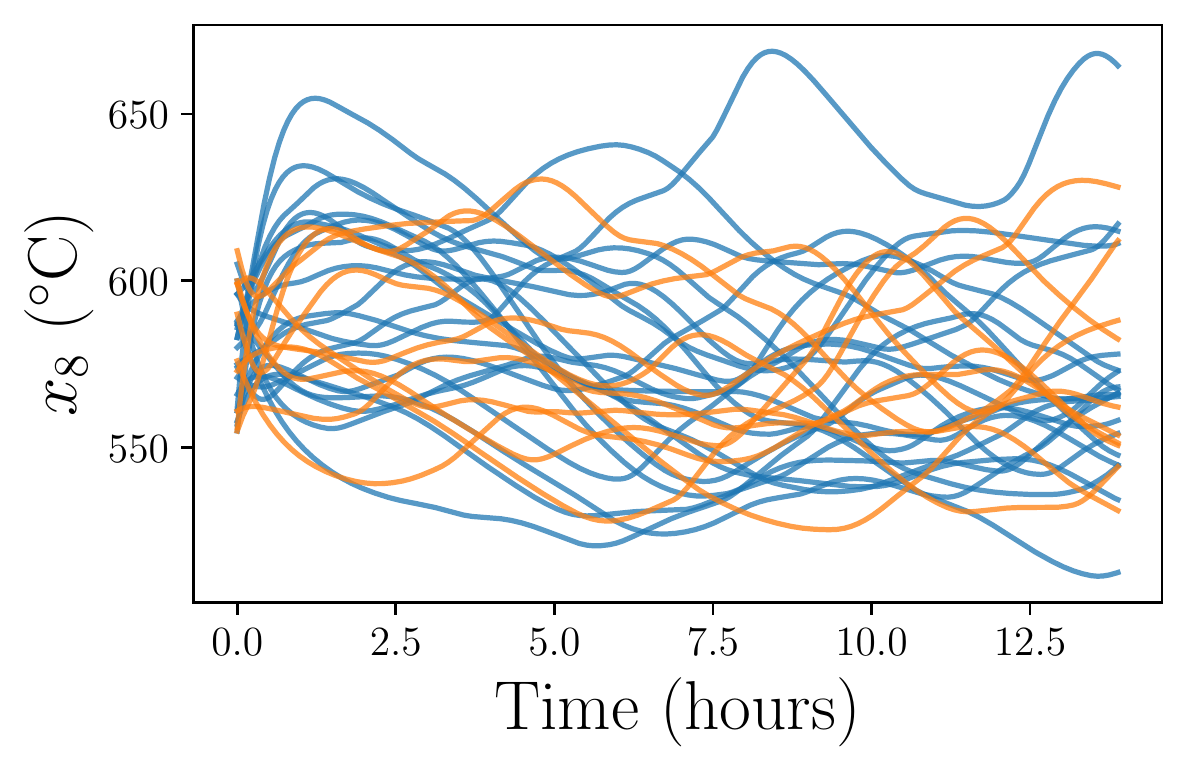}}
        \caption{Side wall temperature $x_8$}
        \label{subfig:training_testset_x8}
    \end{subfigure}\\
    
    \begin{subfigure}[t]{0.32\linewidth}
        \raisebox{-\height}{\includegraphics[width=\linewidth]{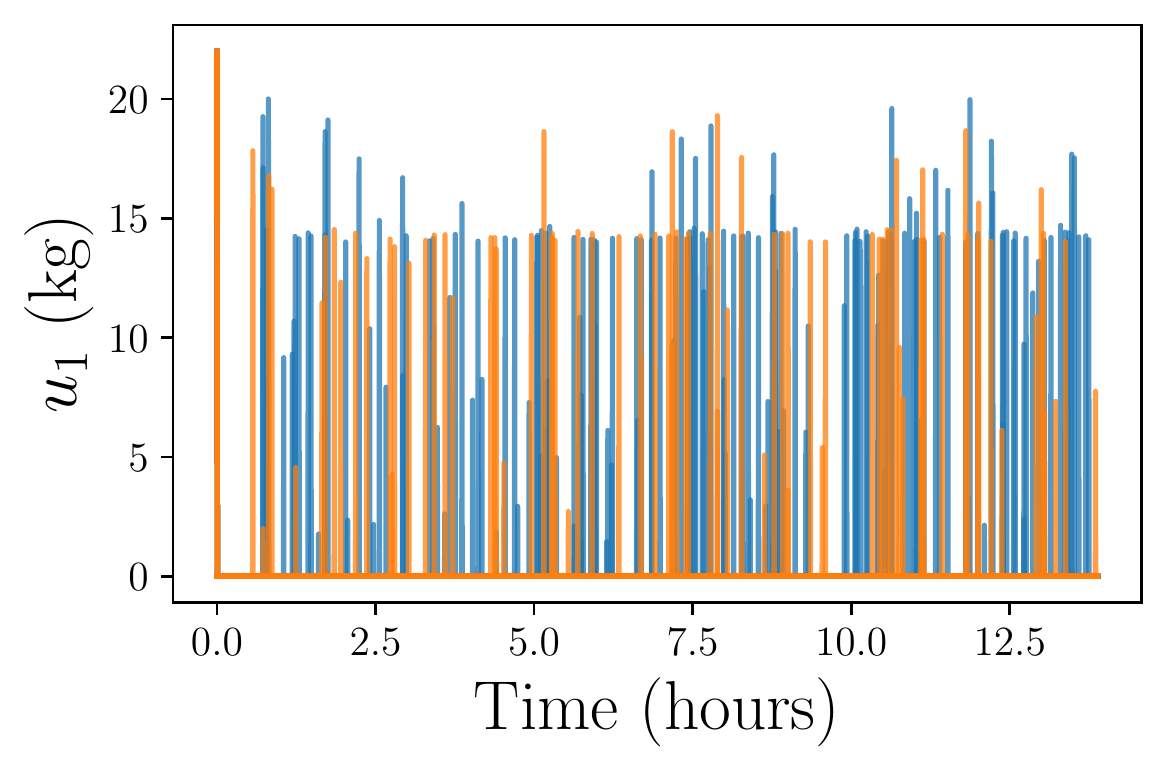}}
        \caption{Alumina feed $u_1$}
        \label{subfig:training_testset_u1}
    \end{subfigure}
        \begin{subfigure}[t]{0.32\linewidth}
        \raisebox{-\height}{\includegraphics[width=\linewidth]{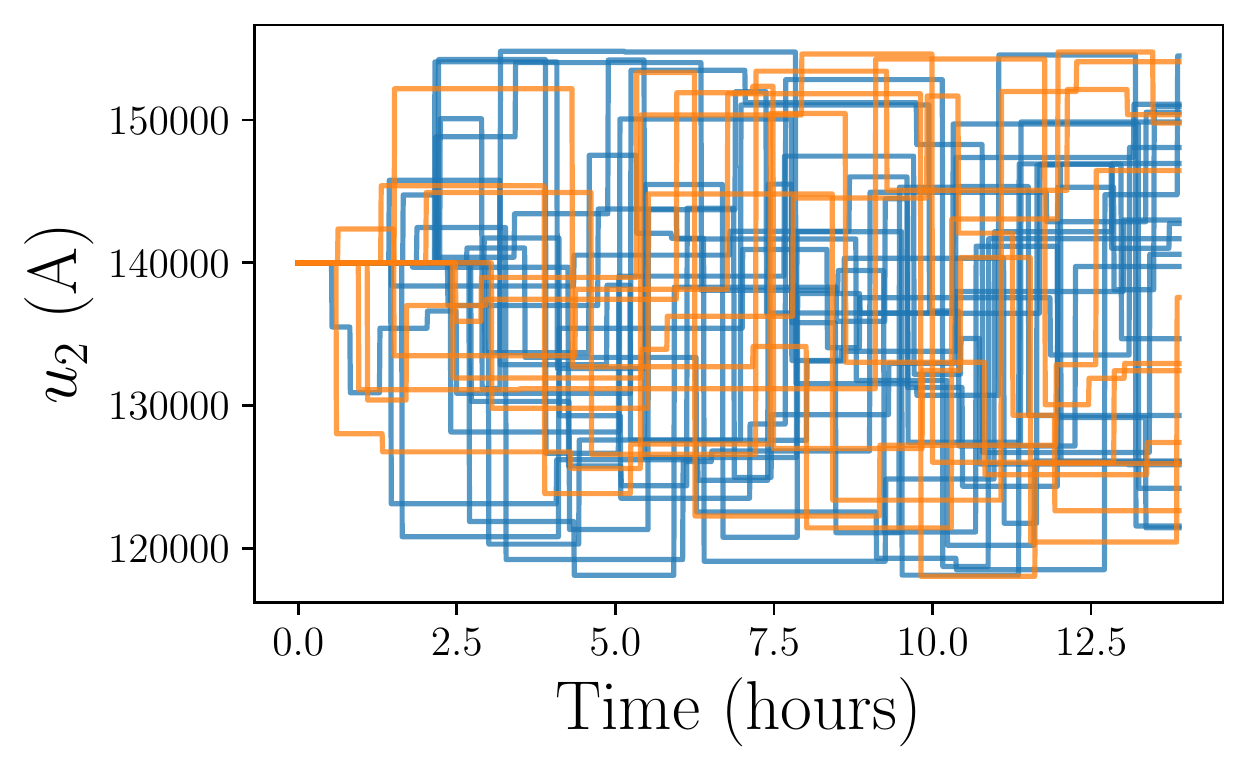}}
        \caption{Line current $u_2$}
        \label{subfig:training_testset_u2}
    \end{subfigure}
        \begin{subfigure}[t]{0.32\linewidth}
        \raisebox{-\height}{\includegraphics[width=\linewidth]{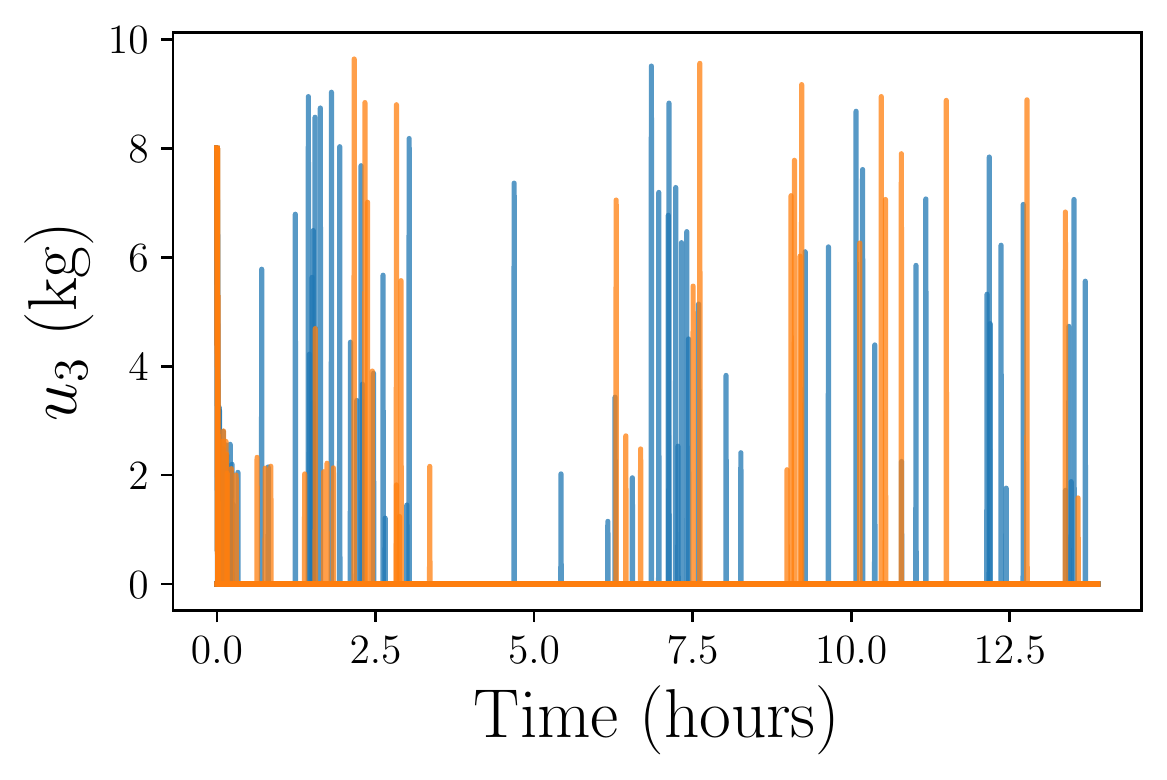}}
        \caption{Aluminum fluoride feed $u_3$}
        \label{subfig:training_testset_u3}
    \end{subfigure}
    \begin{subfigure}[t]{0.32\linewidth}
        \raisebox{-\height}{\includegraphics[width=\linewidth]{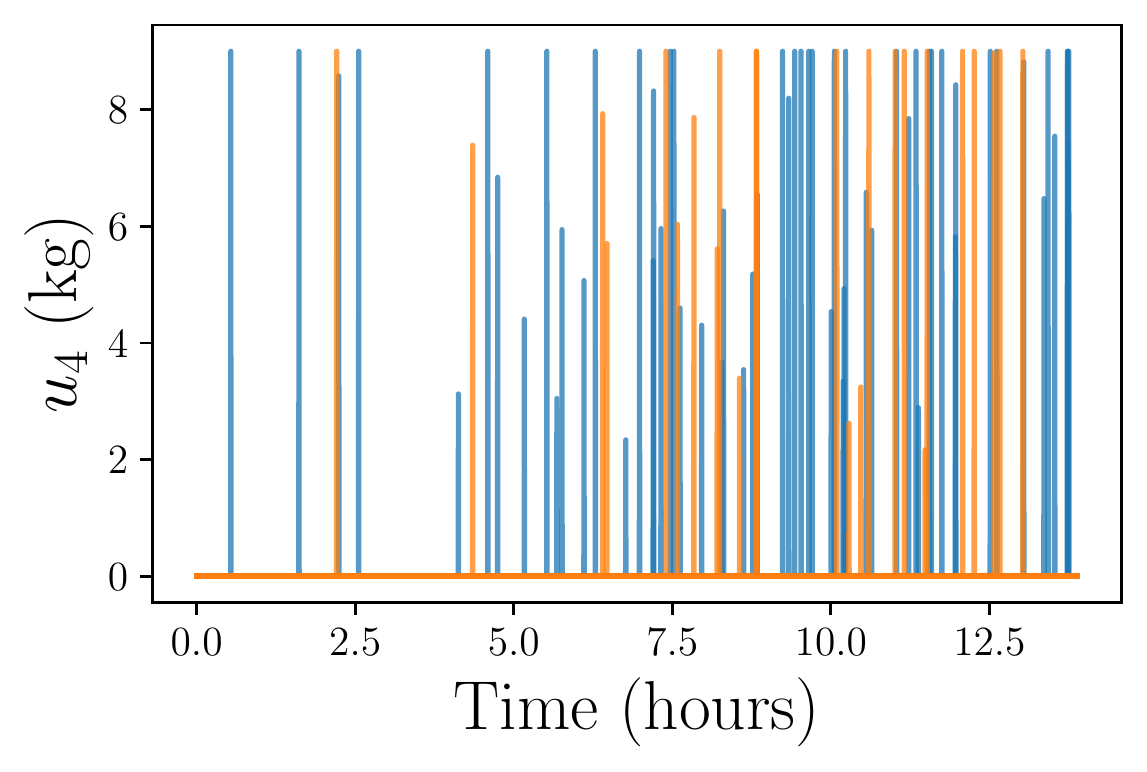}}
        \caption{Metal tapping $u_4$}
        \label{subfig:training_testset_u4}
    \end{subfigure}
    \begin{subfigure}[t]{0.32\linewidth}
        \raisebox{-\height}{\includegraphics[width=\linewidth]{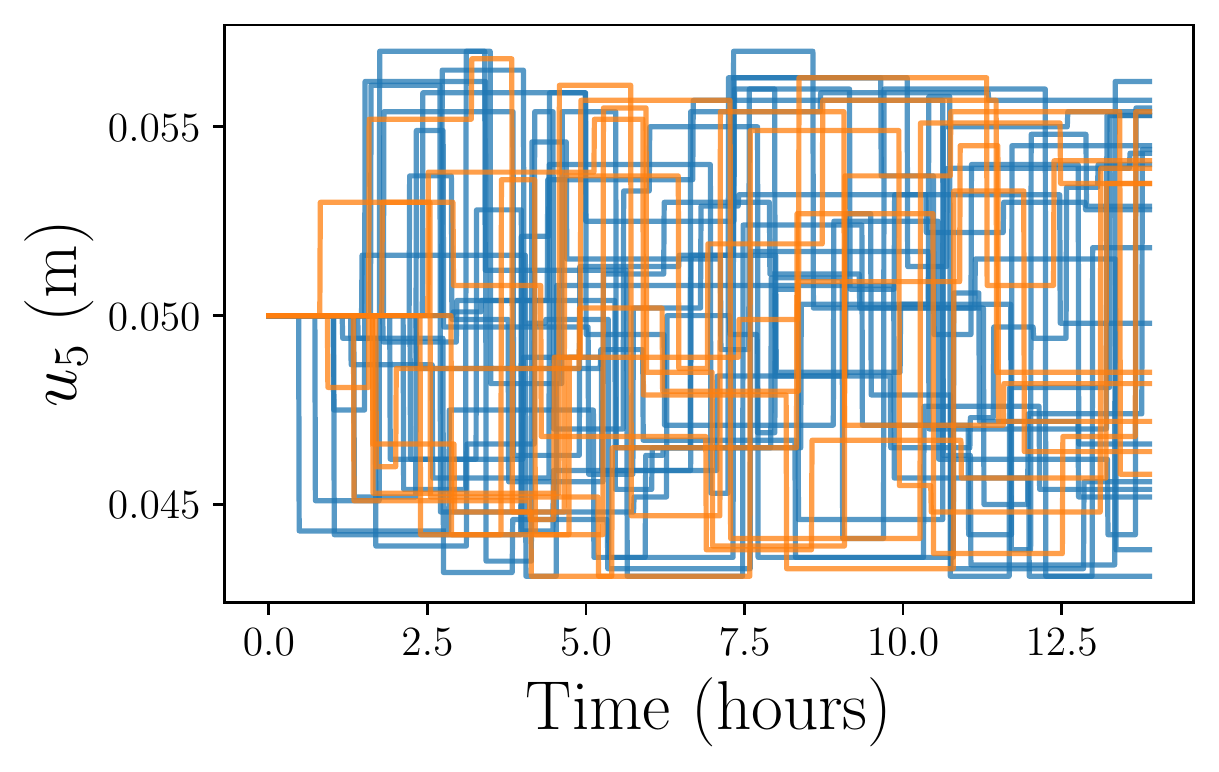}}
        \caption{Anode-cathode distance $u_5$}
        \label{subfig:training_testset_u5}
    \end{subfigure}
        \begin{adjustbox}{max width=\linewidth}
        \begin{tikzpicture}
            \begin{customlegend}[legend columns=2,legend style={draw=none, column sep=1ex},legend entries={Training data, Test data}]
                \addlegendimage{Tblue,thick, sharp plot}
                \addlegendimage{Torange,thick, sharp plot}
            \end{customlegend}
        \end{tikzpicture}
    \end{adjustbox}\\
    \caption{Training and test set trajectories of the system states. Only 10 random sample test trajectories are shown here to make the figures clearer.}
    \label{fig:training_testset}
\end{figure*}

Figure~\ref{fig:training_testset} shows the time series evolution of the entire training set and test set. The training set trajectories are blue while the test set trajectories are orange. The figures show that the range of the training set covers the range of the test set. This indicates that models are evaluated on interpolation cases in the test set.

\subsubsection{Estimation of the regression variable $\mathbf{\dot{x}}$}
The \ac{ODE}s in Equation~\eqref{eq:alu_equations} are time invariant. 
This means that at time $k+1$, $\dot{\state}_{k+1}$ in general only depends on the current state and input $(\state_k, \xinput_k)$ at time $k$. In other words, the system has the Markov property.
Therefore, the datasets are listed in pairs ${\cal D} = \{(\state_k,\mathbf{y}_k)\} = \{(\state_k, \xinput_k), \dot{\state}_k\}$. 
This does not always hold in practice, and the state vector must therefore be augmented with additional information, i.e. lookback states from previous time steps. 
Takens' Theorem gives an upper bound on the number of necessary lookback states \citep{takens_detecting_1981}.
The time derivatives at time $k$ are estimated as the forward difference $\dot{\state}_k = (\state_{k+1} - \state_k)/h$, where $h$ is the time step.
In this work we use $h=\si{10\second}$.
This numerical derivative induces a discretization error. 
However, since the dynamics of the aluminum electrolysis is slow, this error is considered negligible. 
Because the systems are driven by an input signal $\xinput$, we must choose a value for $\xinput_k$ at each time step. 
This choice will greatly affect the variation in the dataset.

\begin{table}[ht]
    \begin{center}\caption{Initial conditions for system variables. For $x_2$ and $x_3$, concentrations $c_{x_2}$ and $c_{x_3}$ are given.}
    \begin{tabular}[width=\textwidth]{l|l}
    \hline
    Variable &  Initial condition interval \\ \hline
    $x_1$ & $[2060,\;4460]$\\
    $c_{x_2}$ & $[0.02,\; 0.05]$ \\
    $c_{x_3}$ & $[0.09,\; 0.13]$  \\
    $x_4$& $[11500,\; 16000]$  \\
    $x_5$ & $[9550,\; 10600]$ \\
    $x_6$ & $[940,\; 990]$ \\
    $x_7$ & $[790,\; 850]$ \\
    $x_8$ & $[555,\; 610]$ \\
    \hline
    \end{tabular}
    \label{table:init_conditions_aluminum}
    \end{center}
\end{table}

\subsubsection{Input signal generation}
While machine learning models are extremely useful for function approximation and interpolating data, they naturally do not always extrapolate properly and are highly dependent on on the quality and variety of the data that they are trained on. 
Due to this it is vital that the training data covers the intended operational space of the system. 
Here, the operational space means the region of the state space which the system operates in, meaning state and input vectors $[\state^T, \xinput^T]^T $ observed over time. 
The data should capture the different nonlinear trends of the system covered by the operational space. 
For systems \textit{without exogenous inputs}, variation can only be induced by simulating the system with different initial conditions $\state(t_0)$.  
For systems \textit{with exogenous inputs}, the initial conditions are generated in the same way. 
Moreover, the input vector $\xinput$ will excite the system dynamics. 
The aluminum process has a feedback controller that ensures safe and prescribed operation. 
However, operational data from a controlled, stable process is generally characterized by a low degree of variation which is insufficient for effective system identification. 
A well-known convergence criterion for the identification of linear time-invariant systems is \ac{PE}.
A signal $\state(t_k)$ is \ac{PE} of order $L$ if all sub-sequences $[\state(t_k),\dots,\state(t_k+L)]$ span the space of all possible sub-sequences of length $L$ that the system is capable of generating.
While the \ac{PE} criterion is not directly applicable to nonlinear systems, sufficient coverage of the dynamics is required for successful system identification \citep{ljung_system_1998, nelles_nonlinear_2020}.

To push the system out of its standard operating conditions, we add random perturbations to the control inputs. In general, each control input $i$ is given by:
\begin{equation}
    u_i = \textrm{Deterministic term + Random term}.
\end{equation}
The control inputs $u_1, \; u_3$ and $u_4$ are impulses. The random term is zero for these control inputs when the deterministic term is zero. The deterministic term is a proportional controller. The control inputs $u_2$ and $u_5$ are always nonzero. These control inputs have constant deterministic terms and a random term that changes periodically.  The random term of the control inputs are determined using the \ac{APRBS} method \citep{WINTER2018802}.
\begin{table}[ht]
    \begin{center}
        \caption{Equations used to control the aluminum process}
        \label{table:control_inputs}
        \begin{adjustbox}{max width=\linewidth}
            \begin{tabular}[width=0.7\textwidth]{l|l|l}
            \hline
            Input & Deterministic term & Random term interval  \\ \hline
            $u_1$ & $3\cdot10^{4}(0.023 - c_{x_2})$ &$[-2.0,2.0]$  \\
            $u_2$ & $1.4\cdot 10^{4}$ &$[-7\cdot10^{3},7\cdot10^{3}]$ \\
            $u_3$ & $1.3\cdot10^{4}(0.105-c_{x_3})$ &$[-0.5, \;0.5]$   \\
            $u_4$& $2(x_5 - 10^{4})$ & $[-2.0, \;2.0]$ \\
            $u_5$ & $0.05$ &$[-0.015, \;0.015]$  \\
            \end{tabular}
        \end{adjustbox}
    \end{center}
\end{table}
Table~\ref{table:control_inputs} gives the numerical values of the deterministic term of the control input, the interval of values for the random terms.

\subsection{Training}
\label{subsec:hyperparameters}

The models were trained on the training set using the total-loss function shown in Equation~\eqref{eq:total-loss}, where the loss function $\loss(\cdot,\cdot)$ is the \ac{MSE} as shown in Equation~\eqref{eq:mse}.
Four different model types were compared:
\begin{itemize}
    \item Dense \ac{NN} 
    \item Sparse \ac{NN} 
    \item PBM + Dense \ac{NN} 
    \item PBM + Sparse \ac{NN} 
\end{itemize}
The dense networks were trained with $\lambda=0$, and sparse networks with $\lambda=10^{-4}$.
The architecture of all networks was $[13,20,20,20,20,8]$ (13 inputs, 8 outputs, 4 hidden layers with 20 neurons each) \todo{How was this chosen?}.
The \texttt{ReLU} activation function was used for all layers except the output layer, which had no activation function.
The same architecture was used for all networks for a fairer comparison. 
All models were trained for 100 epochs (an epoch is defined as one full pass over the dataset).
The ADAM optimiser \citep{kingma_adam_2014} was used with the following default parameters:
Initial learning rate $\eta=10^{3}$,
Gradient forgetting factor $\beta_1 = 0.9$,
and 
Gradient second moment forgetting factor $\beta_2 = 0.999$.

\subsection{Performance metrics}
\label{subsec:performancemetrics}

In this work, we will focus on long-term forecast error as a measure of performance. The initial condition $\state(t_0)$ are given to the models. Then the consecutive $n$ time steps of the states are estimated $\{\mathbf{\hat{x}}(t_1),\;...,\;\mathbf{\hat{x}}(t_n)\}$. This is called a \textit{rolling forecast}. The model estimates the time derivatives of the states $d\mathbf{\hat{x}_i}/dt$ based on the current state $\state(t_i)$ and control inputs $\xinput(t_i)$ and initial conditions $\state_0 = \state(t_0)$, or the estimate of the current state variables $\mathbf{\hat{x}}(t_{i})$ if $t>t_0$:
\begin{equation}
    \frac{d\mathbf{\hat{x}}(t_i)}{dt} = 
    \begin{cases}
    \net(\mathbf{\hat{x}}(t_i), \; \xinput(t_i)),& \text{if } t_i> t_0\\
    \net(\state_0(t_i), \; \xinput(t_i)),& \text{if } t_i = t_0
    \end{cases}
\end{equation}
Then, the next state estimate $\state(t_{i+1})$ is calculated as
\begin{equation}
    \mathbf{\hat{x}}(t_{i+1}) = \mathbf{\hat{x}}(t_i) + \frac{d\mathbf{\hat{x}}(t_i)}{dt}\cdot \Delta T.
\end{equation}
The rolling forecast can be computed for each of the states $x_i$ for one set of test trajectories $\mathcal{S}_{test}$. However, presenting the rolling forecast of multiple test sets would render the interpretation difficult. By introducing a measure called Average Normalized Rolling Forecast Mean Squared Error (AN-RFMSE) that compresses the information about model performance, the models can easily be evaluated on a large number of test sets. The AN-RFMSE is a scalar defined as:
\begin{equation}
    \textrm{AN-RFMSE} = \frac{1}{p}\sum_{i=1}^p\frac{1}{n}\sum_{j=1}^n\left(\frac{\hat{x}_i(t_j) - x_i(t_j)}{std(x_i)}\right)^2,
    \label{eq:AN-RFMSE}
\end{equation}
where $\hat{x}_i(t_j)$ is the model estimate of the simulated state variable $x_i$ at time step $t_j$, $std(x_i)$ is the standard deviation of variable $x_i$ in the training set $\mathcal{S}_{train}$, $p=8$ is the number of state variables and $n$ is the number of time steps the normalized rolling forecast MSE is averaged over. Hence, for every model $\net_j$ and every test set time series $\mathcal{S}_{test}(i)$, there is a corresponding AN-RFMSE.

\section{Results and discussion}
\label{sec:resultsanddiscussion}  

\begin{figure*}
    \centering
    \includegraphics[width=\linewidth]{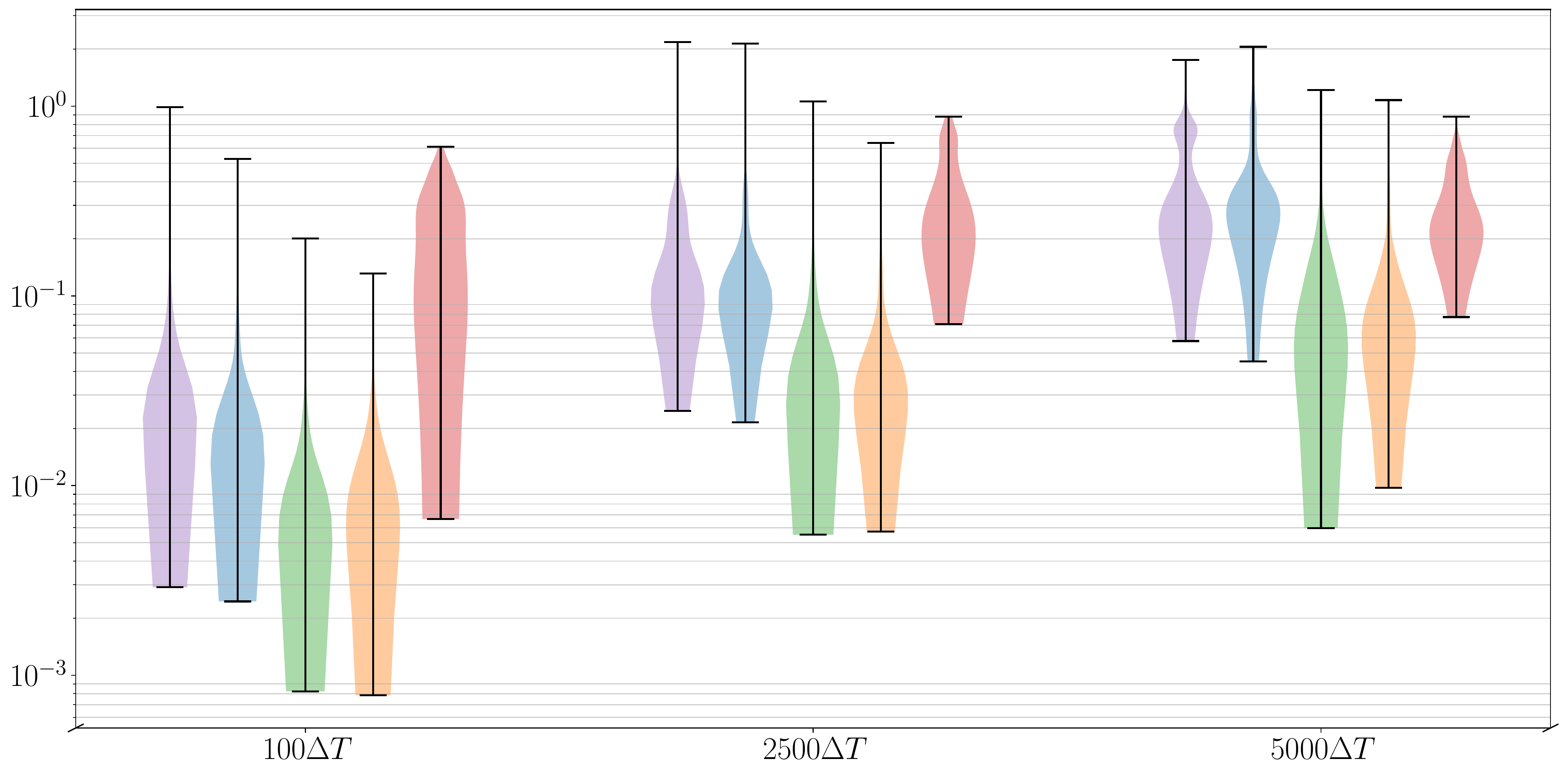}
    \begin{adjustbox}{max width=\linewidth}
        \begin{tikzpicture}
            \begin{customlegend}[legend columns=5,legend style={draw=none, column sep=1ex},legend entries={DDM dense,DDM sparse,CoSTA dense,CoSTA sparse, PBM}]
                \addlegendimage{Tpurple!40, fill=Tpurple!40,area legend}
                \addlegendimage{Tblue!40, fill=Tblue!40,area legend}
                \addlegendimage{Tgreen!40, fill=Tgreen!40,area legend}
                \addlegendimage{Torange!40, fill=Torange!40,area legend}
                \addlegendimage{Tred!40, fill=Tred!40,area legend}
            \end{customlegend}
        \end{tikzpicture}
    \end{adjustbox}
    \caption{Violin-plot of the AN-{RFMSE} for all model types for 100 different initial conditions and inputs signals. The width of the bar reflects the distribution of the data points, and the error bars represent the range of the data. The error is shown after 3 different times to compare the performance in the short, medium, and long-term. We trained 10 different instances for each model type for statistical significance. We see that \ac{CoSTA} improves the predictive accuracy over the whole trajectory. Introducing sparse regularization appears to improve performance for \ac{DDM}, but only appears to affect \ac{CoSTA} models in the long term, where sparse CoSTA appears to have less variance. }
    \label{fig:RFMSE_violin}
\end{figure*}

\begin{figure}
    \centering
    \includegraphics[width=0.95\linewidth]{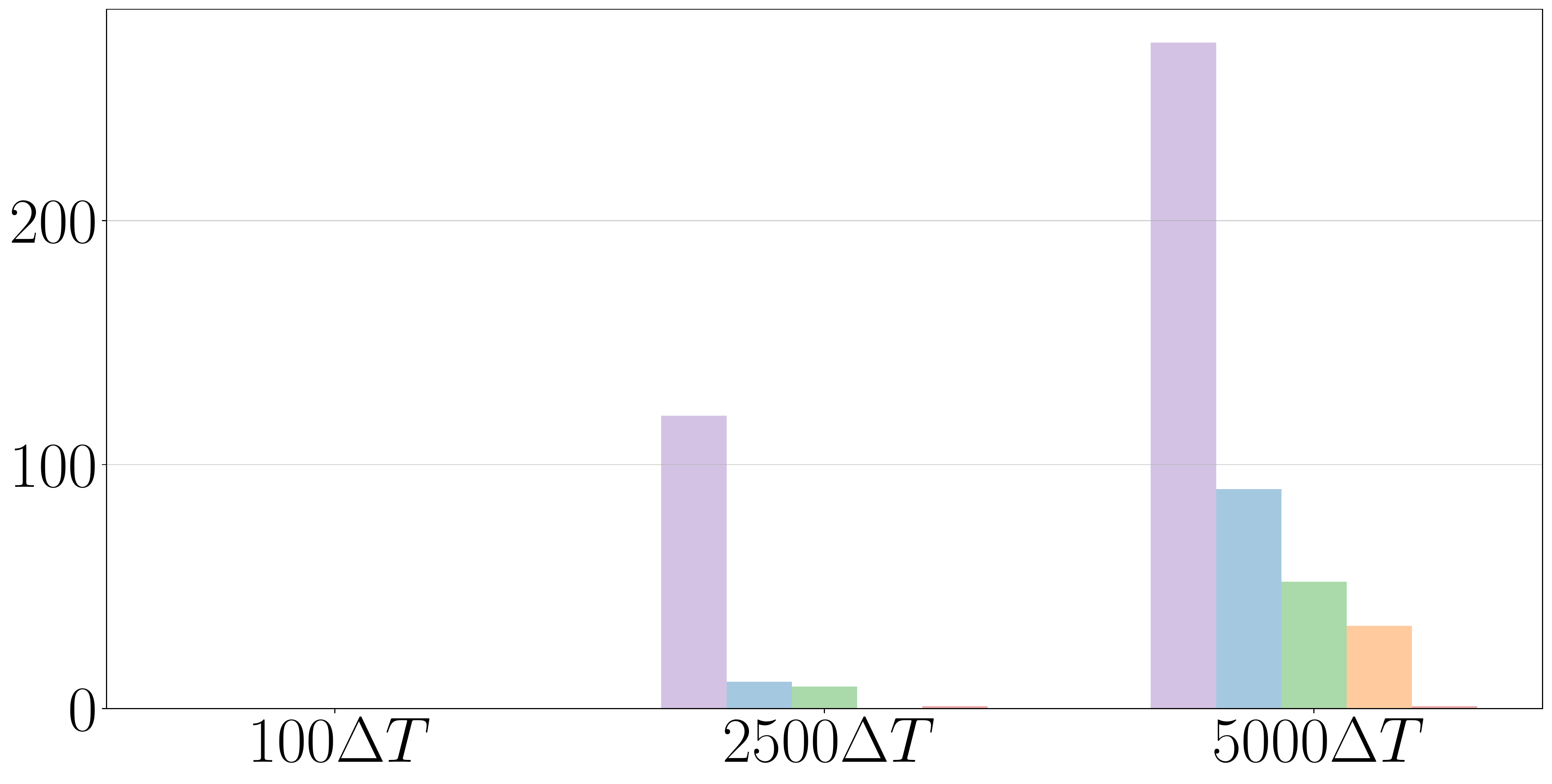}
        \begin{adjustbox}{max width=\linewidth}
        \begin{tikzpicture}
            \begin{customlegend}[legend columns=4,legend style={draw=none, column sep=1ex},legend entries={DDM dense,DDM sparse,CoSTA dense,CoSTA sparse}]
                \addlegendimage{Tpurple!40, fill=Tpurple!40,area legend}
                \addlegendimage{Tblue!40, fill=Tblue!40,area legend}
                \addlegendimage{Tgreen!40, fill=Tgreen!40,area legend}
                \addlegendimage{Torange!40, fill=Torange!40,area legend}
            \end{customlegend}
        \end{tikzpicture}
    \end{adjustbox}
    \caption{Bar chart of the number of times model estimates blow up and diverges. The plot is for all model types for 100 different initial conditions and input signals. The number of blow-ups was counted after 3 different times to compare the performance in the short, medium, and long-term. We trained 10 different instances for each model type for statistical significance. We see that applying \ac{CoSTA} greatly increases the predictive stability in the long term. That is, the number of blow-ups for \ac{CoSTA} models is far less than the number of blow-ups for \ac{DDM}. However, \ac{PBM} does not suffer from significantly fewer blow-ups than \ac{CoSTA}.}
    \label{fig:Divergence_plot}
\end{figure}

For uncertainty quantification, 10 different instances of each of the 4 model types were trained on the same dataset.
Only one instance of the ablated \ac{PBM} was used as defined in Section~\ref{subsec:datageneration}.
All model instances were evaluated on 100 different test trajectories, yielding a total of 4100 data points. 
For some trajectories, some of the model forecasts were found to blow up. 
We set a threshold, where a blow-up is defined as when the normalized \ac{MSE} of the final predicted state exceeds 3.
Figure~\ref{fig:RFMSE_violin} shows a violin plot of the AN-RFMSE for all model types, without the blow-ups. 
The AN-RFMSE is shown at three different times, to demonstrate the short-term, medium-term, and long-term performance of all model types. 
Figure~\ref{fig:Divergence_plot} shows the frequency of blow-ups for each model type.
These results show that on average, all DDM and CoSTA models have a lower RFMSE than the ablated PBM in the short and medium term.
However, we still observe that all DDM and CoSTA models experience some blow-ups in the long term, which the PBM model does not. 
The dense DDM fared the worst, as $27.3\%$ of the forecasts were found to result in blow-ups in the long term. 
The sparse DDM marginally improves on the RFMSE, but we found that the blow-up rate was greatly reduced in the long term compared to the dense DDM. 
Both dense and sparse \ac{CoSTA} models were found to be significantly more accurate than the DDM models. 
The sparse \ac{CoSTA} had similar accuracy to the dense \ac{CoSTA} models in the short and medium term. 
However, the sparse \ac{CoSTA} model had no blow-ups in the short and medium term and had half the blow-up rate of the Sparse \ac{DDM} in the long term. 
These experiments demonstrate that \ac{CoSTA} can reliably correct misspecified \ac{PBM}s and that it greatly improves predictive stability in comparison to end-to-end learning.
The base \ac{PBM} does not exhibit any blow-up issues, suggesting that the blow-ups can be attributed to the \ac{NN}s used in this work.
If long-term forecasts are required ($>3000$ timesteps), we recommend combining the \ac{CoSTA} approach with a sanity check mechanism to detect potential blow-ups.

\begin{figure*}
     \centering
    \begin{subfigure}[t]{0.32\linewidth}
        \raisebox{-\height}{\includegraphics[width=\linewidth]{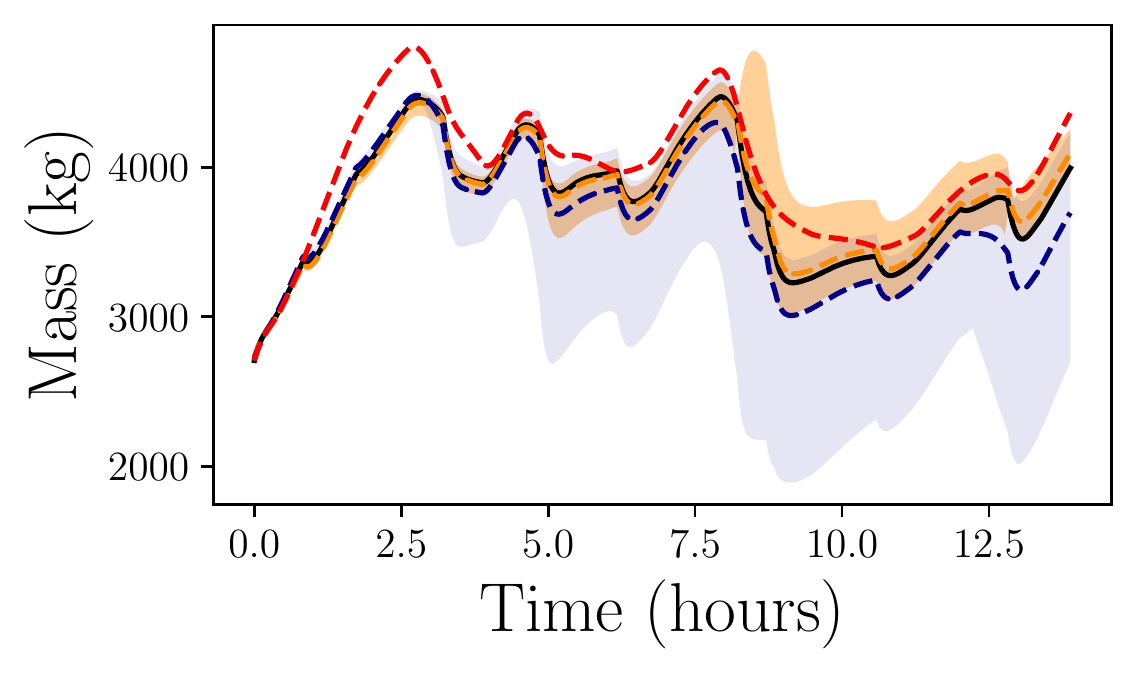}}
        \caption{Side ledge mass $x_1$}
        \label{subfig:Rolling_forecast_x1}
    \end{subfigure}
    \begin{subfigure}[t]{0.32\linewidth}
        \raisebox{-\height}{\includegraphics[width=\linewidth]{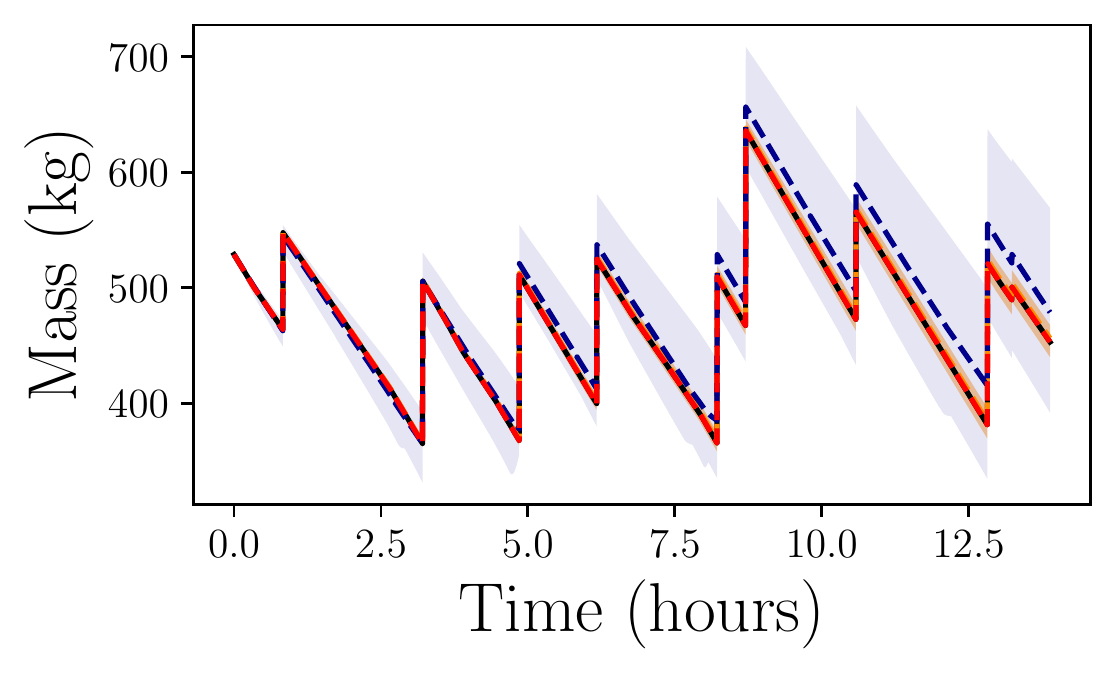}}
        \caption{Alumina mass $x_2$}
        \label{subfig:Rolling_forecast_x2}
    \end{subfigure}
    \begin{subfigure}[t]{0.32\linewidth}
        \raisebox{-\height}{\includegraphics[width=\linewidth]{figures/Rolling_forecast_x2.pdf}}
        \caption{Aluminum fluoride mass $x_3$}
        \label{subfig:Rolling_forecast_x3}
    \end{subfigure}
    \begin{subfigure}[t]{0.32\linewidth}
        \raisebox{-\height}{\includegraphics[width=\linewidth]{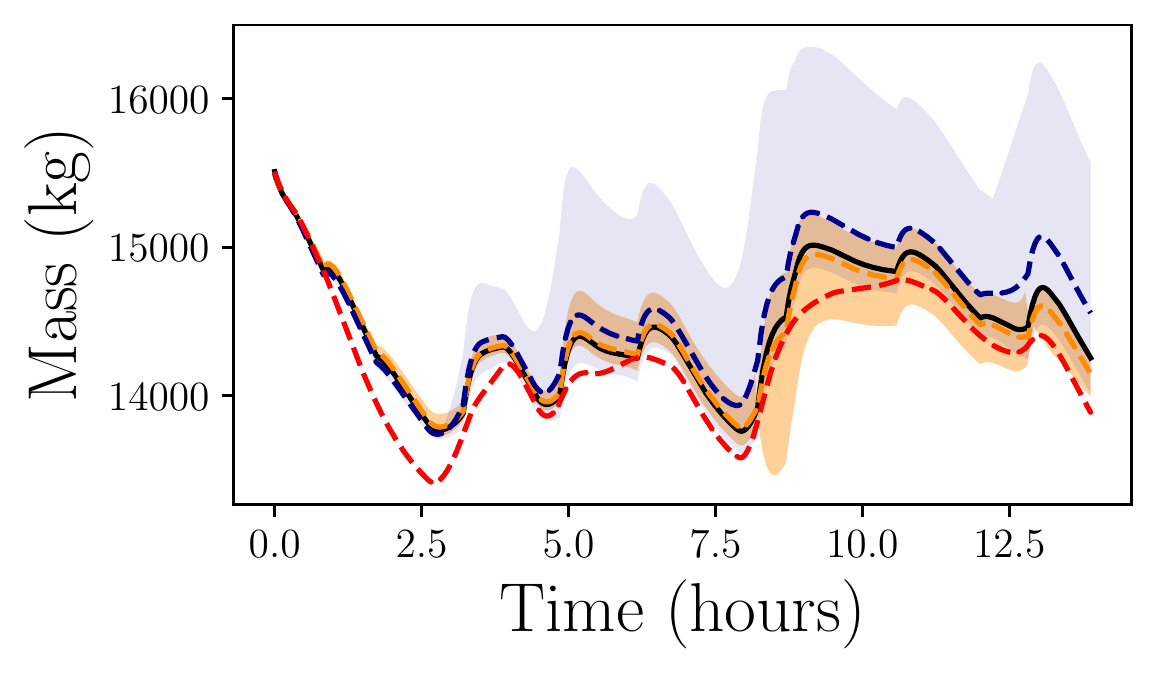}}
        \caption{Molten cryolite mass $x_4$}
        \label{subfig:Rolling_forecast_x4}
    \end{subfigure}
    \begin{subfigure}[t]{0.32\linewidth}
        \raisebox{-\height}{\includegraphics[width=\linewidth]{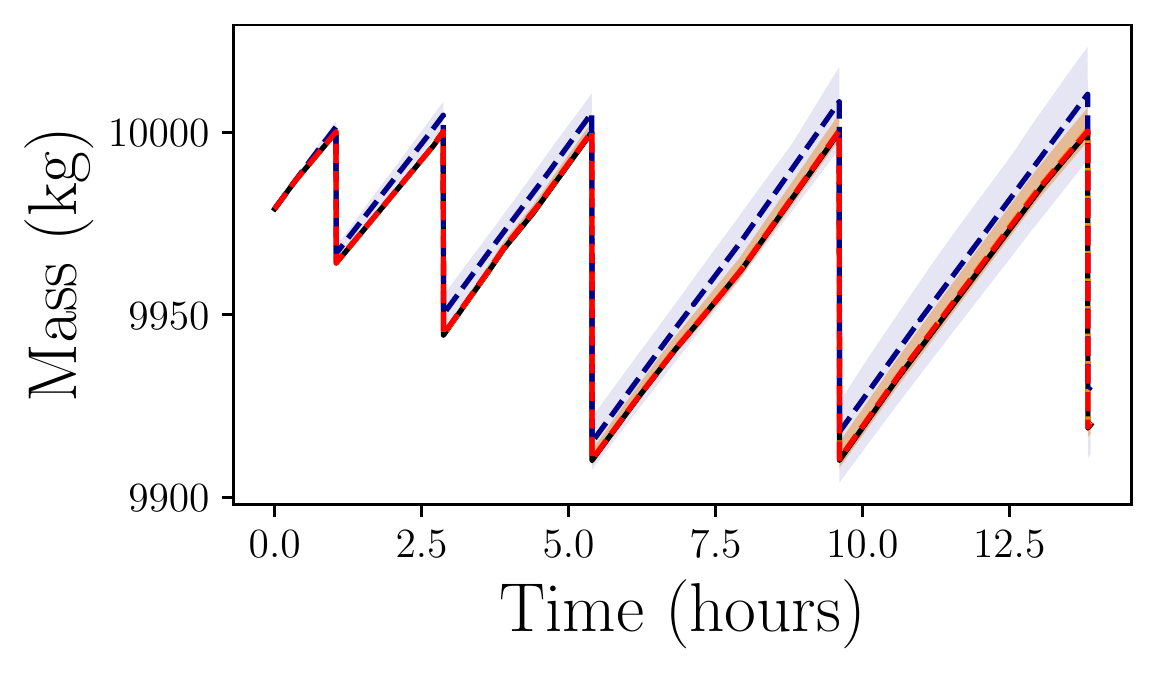}}
        \caption{Produced aluminum mass $x_5$}
        \label{subfig:Rolling_forecast_x5}
    \end{subfigure}
    \begin{subfigure}[t]{0.32\linewidth}
        \raisebox{-\height}{\includegraphics[width=\linewidth]{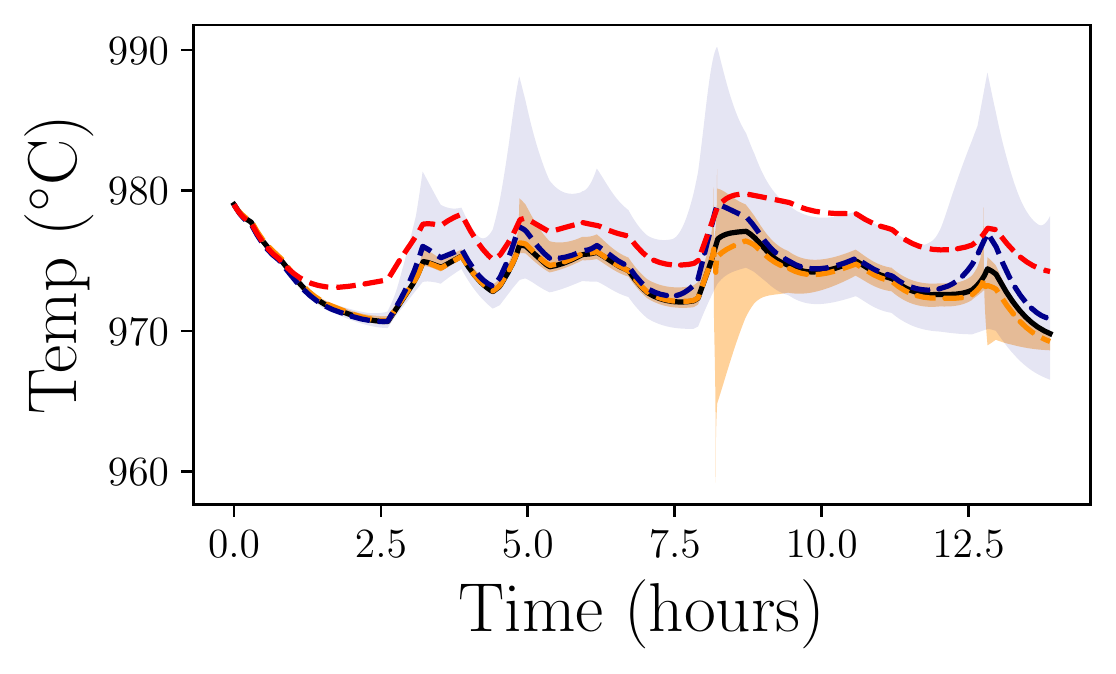}}
        \caption{Bath temperature $x_6$}
        \label{subfig:Rolling_forecast_x6}
    \end{subfigure}
    \begin{subfigure}[t]{0.32\linewidth}
        \raisebox{-\height}{\includegraphics[width=\linewidth]{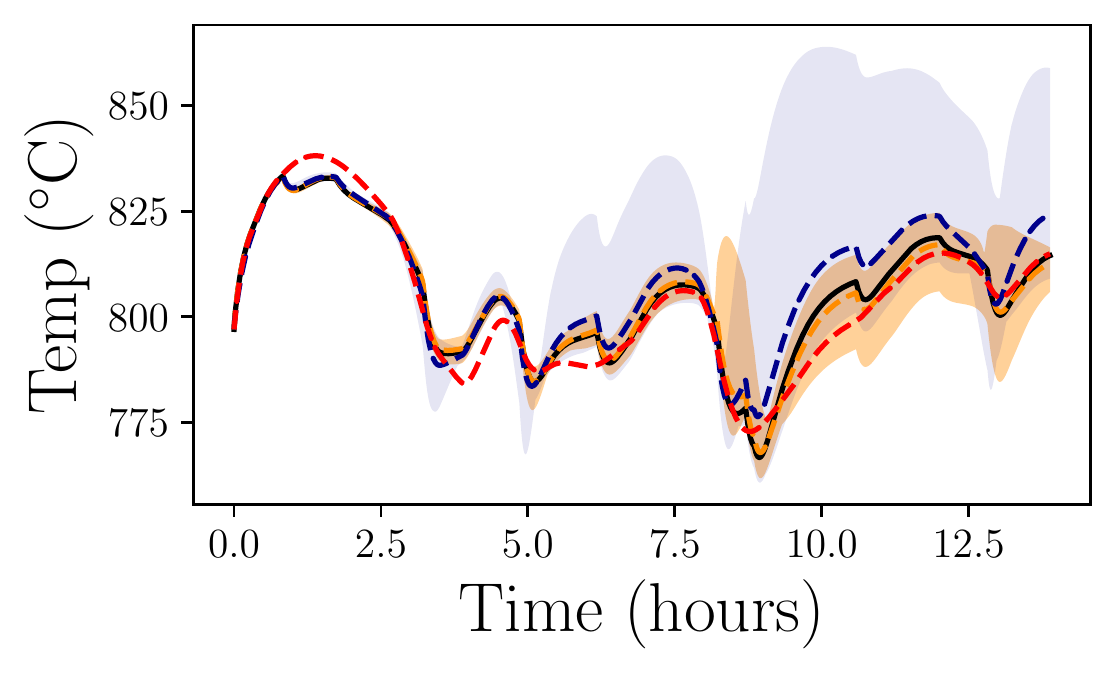}}
        \caption{Side ledge temperature $x_7$}
        \label{subfig:Rolling_forecast_x7}
    \end{subfigure}
    \begin{subfigure}[t]{0.32\linewidth}
        \raisebox{-\height}{\includegraphics[width=\linewidth]{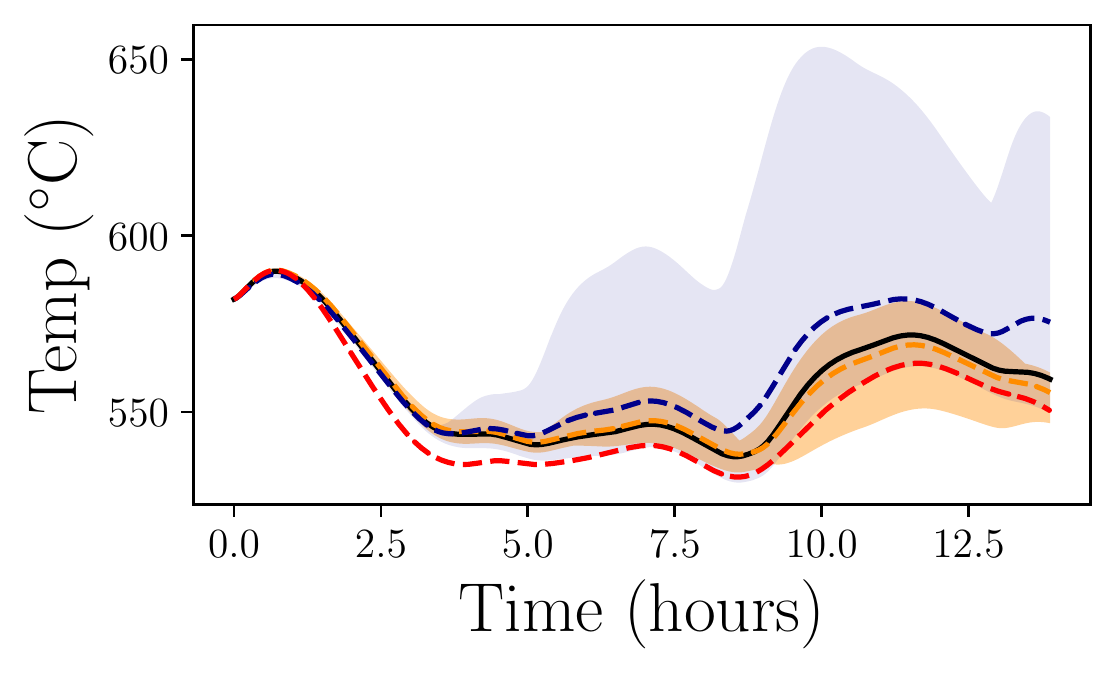}}
        \caption{Side wall temperature $x_8$}
        \label{subfig:Rolling_forecast_x8}
    \end{subfigure}
    \begin{subfigure}[t]{0.32\linewidth}
        \raisebox{-\height}{\includegraphics[width=\linewidth]{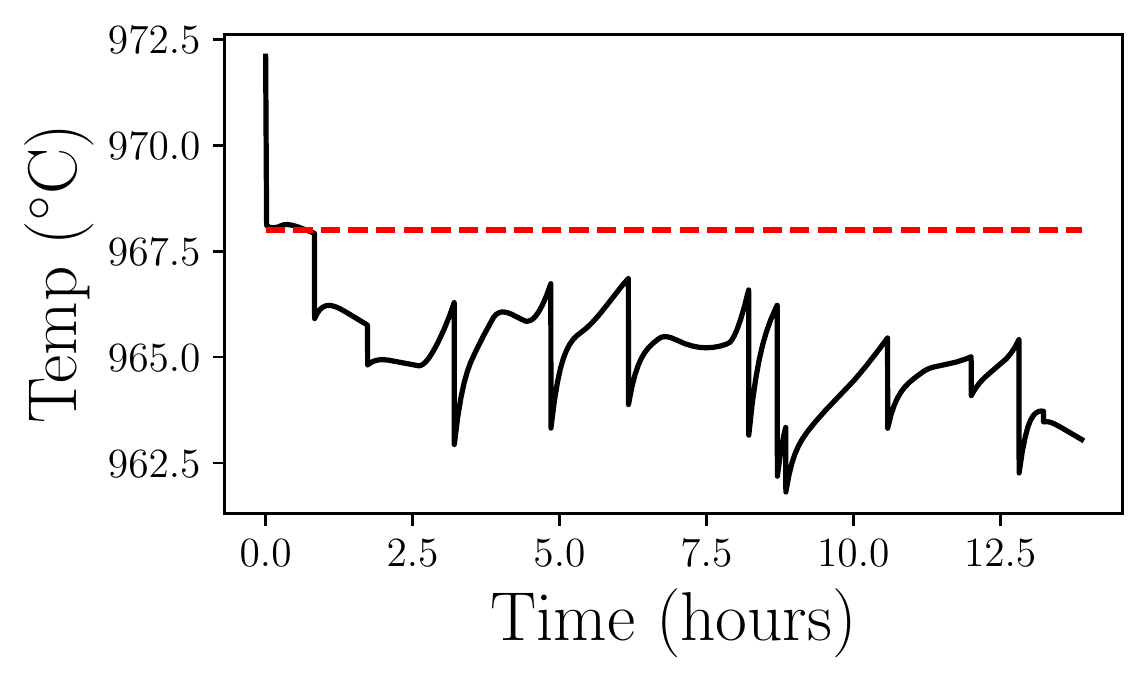}}
        \caption{Liquidus temperature $g_1$}
        \label{subfig:Rolling_forecast_g1}
    \end{subfigure}\\
    \begin{subfigure}[t]{0.32\linewidth}
        \raisebox{-\height}{\includegraphics[width=\linewidth]{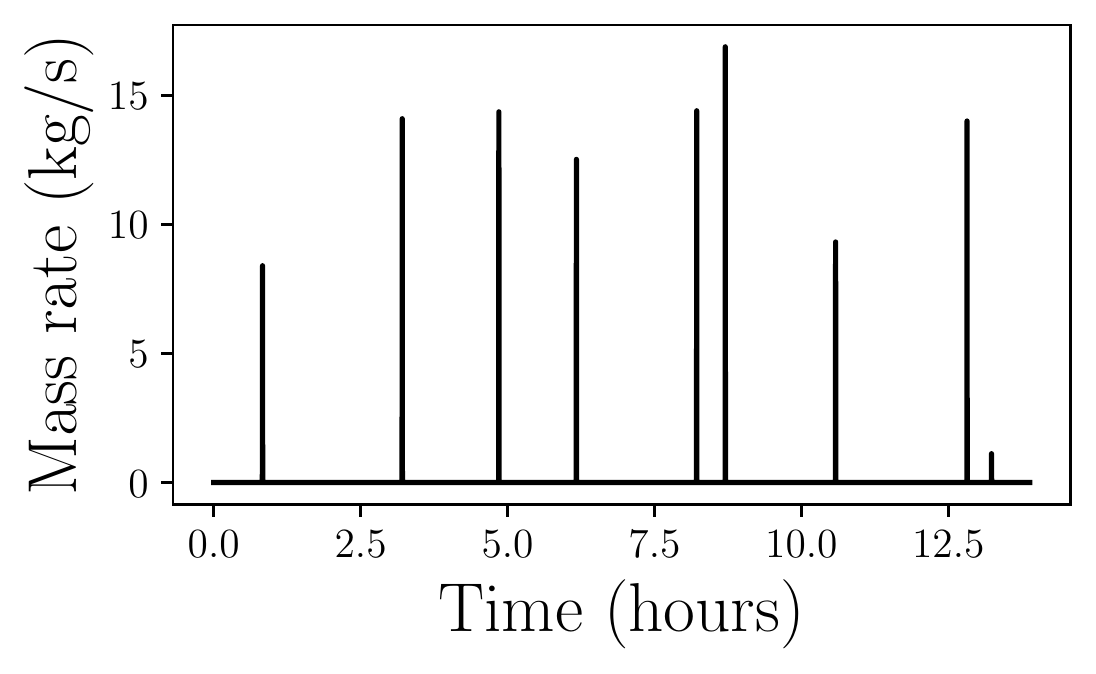}}
        \caption{Alumina feed $u_1$}
        \label{subfig:Rolling_forecast_u1}
    \end{subfigure}
        \begin{subfigure}[t]{0.32\linewidth}
        \raisebox{-\height}{\includegraphics[width=\linewidth]{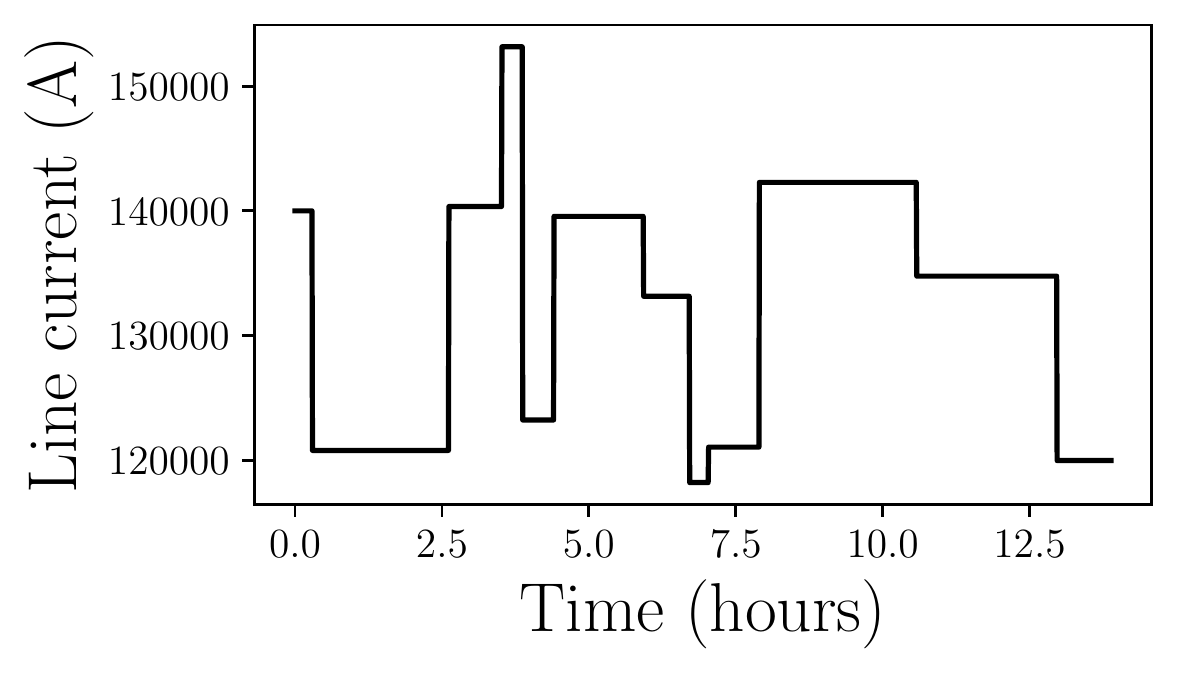}}
        \caption{Line current $u_2$}
        \label{subfig:Rolling_forecast_u2}
    \end{subfigure}
        \begin{subfigure}[t]{0.32\linewidth}
        \raisebox{-\height}{\includegraphics[width=\linewidth]{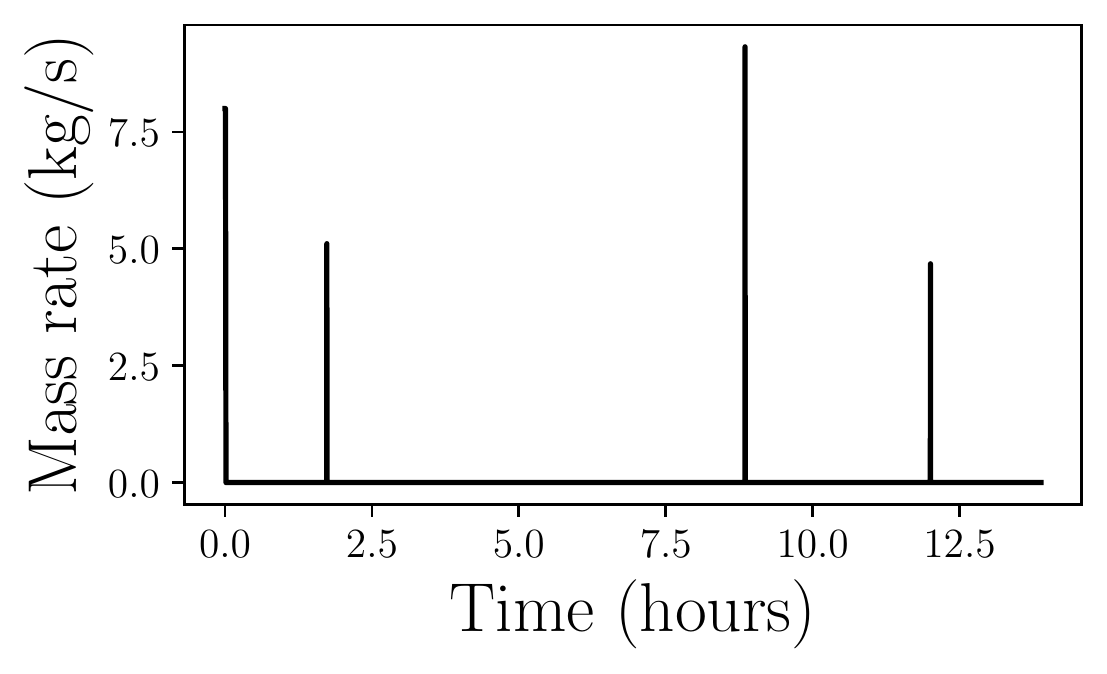}}
        \caption{Aluminum fluoride feed $u_3$}
        \label{subfig:Rolling_forecast_u3}
    \end{subfigure}
    \begin{subfigure}[t]{0.32\linewidth}
        \raisebox{-\height}{\includegraphics[width=\linewidth]{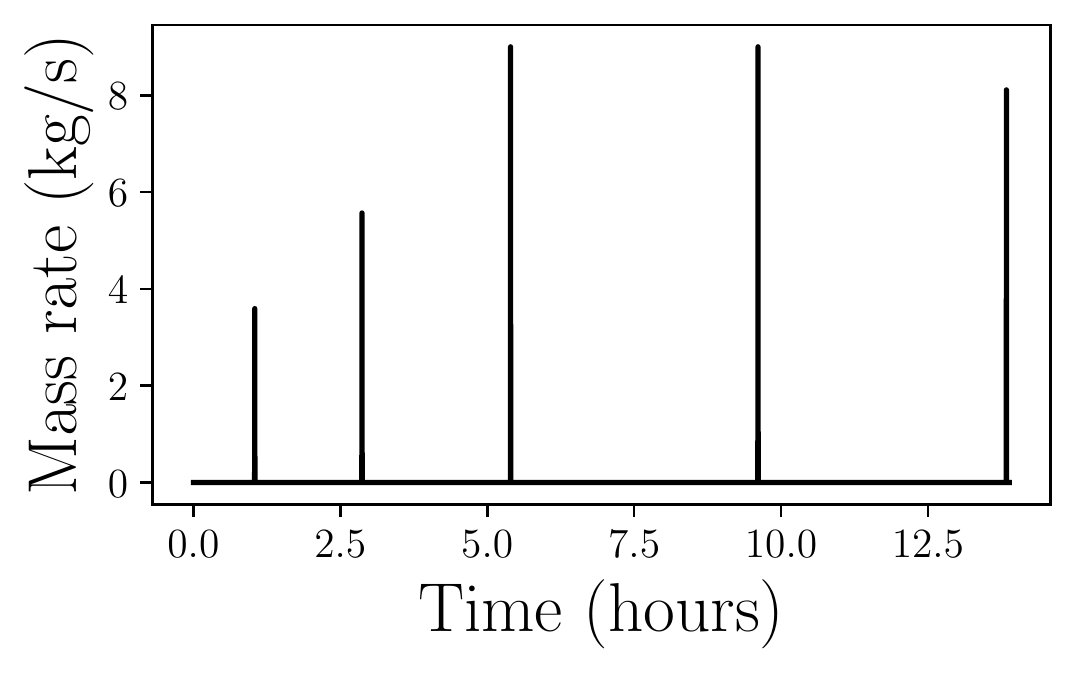}}
        \caption{Metal tapping $u_4$}
        \label{subfig:Rolling_forecast_u4}
    \end{subfigure}
    \begin{subfigure}[t]{0.32\linewidth}
        \raisebox{-\height}{\includegraphics[width=\linewidth]{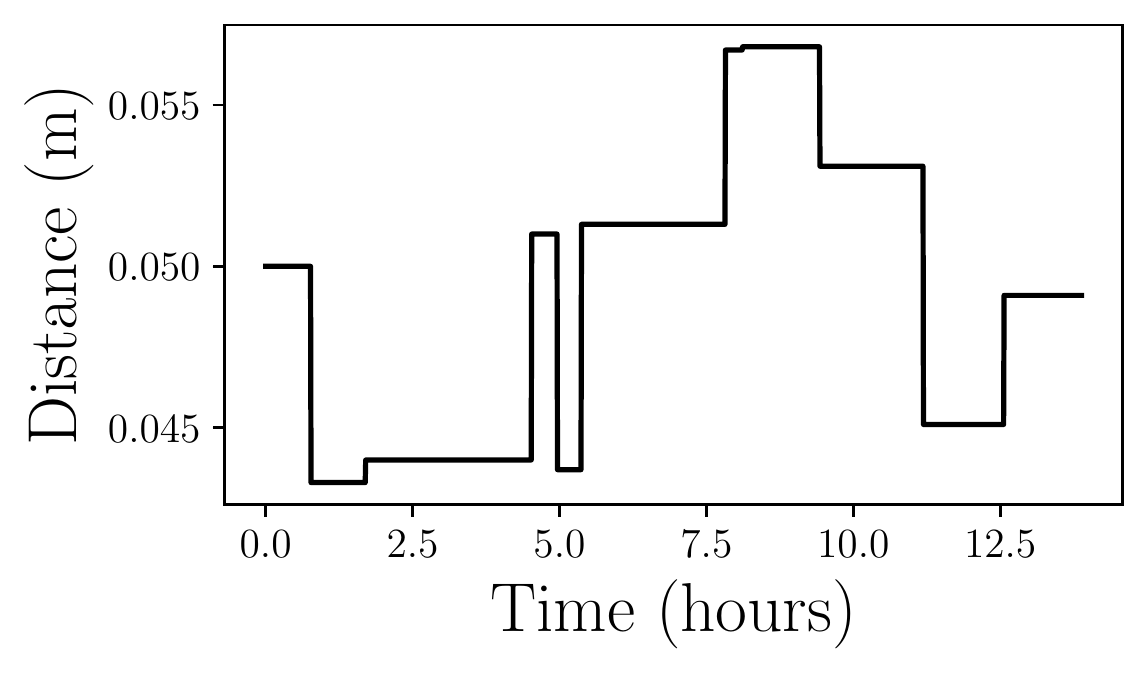}}
        \caption{Anode-cathode distance $u_5$}
        \label{subfig:Rolling_forecast_u5}
    \end{subfigure}
        \begin{adjustbox}{max width=1\linewidth}
        \begin{tikzpicture}
            \begin{customlegend}[legend columns=6,legend style={draw=none, column sep=2ex},legend entries={Truth,CoSTA sparse,DDM sparse,PBM,$99.7\%$ conf. DDM,$99.7\%$ conf. CoSTA}]
                \addlegendimage{black,thick, sharp plot}
                \addlegendimage{Torange,thick, dashed,sharp plot}
                \addlegendimage{Tblue,thick, dashed,sharp plot}
                \addlegendimage{Tred,thick, dashed,sharp plot}
                \addlegendimage{Tpurple!30, fill=Tpurple!20, area legend}
                \addlegendimage{Torange!40, fill=Torange!40, area legend}
            \end{customlegend}
        \end{tikzpicture}
    \end{adjustbox}
    \caption{Rolling forecast of a representative test trajectory. 10 \ac{CoSTA} models with sparse corrective \ac{NN}'s, 10 \ac{DDM}'s consisting of sparse \ac{NN} models, as well as a \ac{PBM}, are predicting the test set trajectories given the initial conditions and the input vector at any given time.}
    \label{fig:Rolling_forecast}
\end{figure*}

Figure~\ref{fig:Rolling_forecast} shows the mean predictions for each model type for a representative test trajectory, along with a $99.7\%$ confidence interval to show the spread of the predictions from the 10 instances of each model type. For better clarity, only the sparse models are shown, due to their superior performance compared to their dense counterparts. Before discussing the differences between the models, we will describe the dynamics of the system, and how the incorrect \ac{PBM} behaves in comparison.

First, note that all variables are non-negative, as they reflect different physical quantities in the system, i.e. mass, temperature, and current.
Inspecting Equation~\eqref{eq:alu_equations}, we see that the states $x_2$, $x_3$, and $x_5$ are linearly dependent on $u_1$, $u_2$, $u_3$, and $u_4$.
We refer to these as the \textit{linear states}, and the rest as the \textit{nonlinear states}.

\todo{
\begin{itemize}
    \item Why is the true system the way it is
    \item What are the differences between the PBM and truth? Why?
    \item Try to explain why DDM succeeds/fails
    \item Try to explain why CoSTA succeeds/fails
    \item From the perspective of the inputs "$u_1$ spikes, so this happens")
\end{itemize}
}

\paragraph{Liquidus temperature $g_1$:}
Figure~\ref{subfig:Rolling_forecast_g1} shows the true liquidus temperature $g_1$ (in black) and the constant \ac{PBM} estimate of the liquidus temperature (in red dotted line). The liquidus temperature $g_1$, which is the temperature at which the bath solidifies, is determined by the chemical composition of the bath. That is, $g_1$ is determined by the mass ratios between $x_2$, $x_3$, and $x_4$. The fact that \ac{PBM} assumes $g_1$ to be constant induces modeling errors for the \ac{PBM}.

\paragraph{Mass of side ledge $x_1$:} Figure~\ref{subfig:Rolling_forecast_x1} shows the mass of frozen cryolite (\ce{Na_3 Al F_6}), or side ledge. The solidification rate $\dot{x}_1$ is proportional to the heat transfer $Q_{liq-sl}$ through the side ledge ($Q_{liq-sl}\sim \left( \frac{g_1 - x_7}{x_1}\right)$) minus the heat transfer $Q_{bath-liq}$ between the side ledge and the bath ($Q_{bath-liq}\sim (x_6 - g_1)$). 
The solidification rate $\dot{x}_1$ is dependent on the value of $g_1$, and therefore the \ac{PBM} incorrectly predicts the mass rate $\dot{x}_1$. In Figure~\ref{subfig:Rolling_forecast_x1} we see that the \ac{PBM} modeling error for $x_1$ starts to increase after approximately one hour. This is simultaneously as the true liquidus temperature $g_1$ drifts away from the constant \ac{PBM} estimate of $g_1$, see Figure~\ref{subfig:Rolling_forecast_g1}. As we can see, the \ac{PBM} overestimates $g_1$. Therefore, the \ac{PBM} also overestimates the heat transfer out of the side ledge, leading to an overestimate of the amount of cryolite that freezes, and hence an overestimate of the increase in side ledge mass. However, this modeling error is limited by the effect that an increased side ledge mass (and therefore increased side ledge thickness) leads to better isolation. Thus, the \ac{PBM} estimate of the heat transfer through the side ledge $Q_{liq-sl}$ is inversely proportional to the $x_1$ estimate, and the modeling error of $x_1$ reaches a steady state for a constant modeling error in $g_1$. In addition to modeling errors due to errors in the $g_1$ estimate, modeling errors of $x_6$ and $x_7$ propagates as modeling errors in $\dot{x_1}$. 

Both the mean of DDM and the mean of CoSTA models appear to correctly predict the response of $x_1$. However, both model classes show a growing spread.
While the error spread of both model classes appears to grow over time, the DDM error grows roughly twice as fast. Furthermore, both CoSTA and DDM show some cases where the error bound becomes significantly large, meaning that one or more of the models fail. For the DDM models, these cases appear more frequently, and the errors are larger than for the CoSTA models. Figures~\ref{subfig:Rolling_forecast_x6} and \ref{subfig:Rolling_forecast_x7} shows that these error peaks often coincide with the peaks in the bath temperature $x_6$ and the side ledge temperature $x_7$.

\paragraph{Mass of alumina $x_2$:}
Figure~\ref{subfig:Rolling_forecast_x2} shows the mass of aluminum in the bath. Equation~\eqref{eq:alu_equations} shows that $\dot{x}_2$ (mass rate of \ce{Al2O3}) is proportional to $u_1$ (\ce{Al2O3} feed), and negatively proportional to $u_2$. Figure~\ref{subfig:Rolling_forecast_x2} shows that this yields a saw-tooth response that rises as $u_1$ spikes, and decays with a rate determined by $u_2$. This state has no dependence on $g_1$, nor any dependence on other states that depends on $g_1$. Therefore the PBM (and CoSTA) predict this state with no error. On the other hand, the spread of the DDM models grows over time, with the mean error eventually becoming significant. 

\paragraph{Mass of aluminum fluoride $x_3$:}
The $x_3$ state (mass of \ce{Al F_3}) acts as an accumulator, rising when \ce{Al F_3} is added to the process ($u_3$ spikes), and falling when \ce{Al2O3} is added to the process ($u_1$ spikes). The latter is caused by impurities (\ce{Na_2 O}) in the Alumina (\ce{Al2 O3}) reacting with \ce{Al F_3}, generating cryolite (\ce{Na_3 Al F_6}). 
As can be seen in Figure~\ref{subfig:Rolling_forecast_x3}, the latter effect is relatively small.
Despite this, the DDM appears to correctly model these decreases.
However, the DDM models become less and less accurate as time goes on.
The PBM and CoSTA model $x_3$ with no error. 

\paragraph{Mass of molten cryolite $x_4$:}
This state represents the mass of molten cryolite in the bath, where $\dot{x}_4 = k_5 u_1 -\dot{x}_1$. 
The first term represents additional cryolite generated by reactions between impurities in the added alumina ($u_1$) and \ce{AlF_3} ($x_3$). 
The second term describes how the cryolite can freeze ($x_1$) on the side ledge, which can melt again as the side-ledge temperature $x_7$ increases.
As can be seen in Figure~\ref{subfig:Rolling_forecast_x4} the response of $x_4$ therefore mirrors that of $x_1$, with relatively small upturns when alumina is added ($u_1$). 
Inspecting Figure~\ref{fig:Rolling_forecast}a, we see that the models have essentially identical behavior.
Incorrectly estimating $x_4$ causes some issues.
The mass ratio $c_{x_2}$ (see Equation~\eqref{eq:alu_ratios}) is important in terms of determining the cell voltage $U_{cell}$. A forecasting error of $x_4$ will propagate as a forecasting error of $c_{x_2}$, leading to inaccurate estimates of the cell voltage $U_{cell}$. This is elaborated when discussing the bath temperature $x_6$.

\paragraph{Mass of produced metal $x_5$:}
This linear state also has a saw-tooth characteristic, growing at a rate proportional to the line current ($u_2$), and falling when metal is tapped ($u_4$ spikes).
Looking at Figure~\ref{subfig:Rolling_forecast_x5}, the DDM models have similar error dynamics to the other linear states, while the PBM and CoSTA models have virtually no error.

\paragraph{Temperature in the bath $x_6$:}
There are several possible sources of \ac{PBM} modeling errors of the bath temperature $x_6$. As discussed earlier, since the \ac{PBM} overestimates the side ledge thickness due to a modeling error of $g_1$, it follows that the \ac{PBM} overestimates the thermal insulation of the side ledge. This leads to an overestimation of the bath temperature, as the heat transfer out of the bath is underestimated. 
In Figure~\ref{subfig:Rolling_forecast_x6}, we see this overestimate of $x_6$ provided by the \ac{PBM} after approximately one hour, simultaneously as the \ac{PBM} starts to overestimate the side ledge mass $x_1$.

Furthermore, the change in bath temperature $\dot{x}_6$ is determined by the energy balance in the bath. The energy balance in the bath consists of several components, namely the electrochemical power $P_{el}$ which adds energy to the system, the heat transfer from the bath to the side ledge $Q_{bath-sl}$ which transports energy out of the bath, and the energy  $E_{tc, liq}$ required to break inter-particle forces in the frozen cryolite liquidus temperature. The electrochemical power $P_{el} = U_{cell}\cdot u_2$ is the product of the cell voltage $U_{cell}$ and the line current $u_2$. The cell voltage is given by $U_{cell} =  \left( g_5 + \frac{u_2 u_5}{2620 g_2}\right)$, where $g_5$ is the bubble voltage drop, and $\frac{u_2 u_5}{2620 g_2}$ is the voltage drop due to electrical resistance in the bath. The bubble voltage drop $g_5$ increases exponentially when the operation gets close to an anode effect. Anode effects occurs when the mass ratio of alumina - $c_{x_2}$ is reduced to the a critical mass ratio of alumina $c_{x_2,crit} \sim 2$.  This can explain overestimate error peaks in the $x_6$ estimate, which are most present for the \ac{DDM} models. As we can see in Figure~\ref{subfig:Rolling_forecast_x6} the peaks of the error band for the DDM happens simultaneously with overestimates of $x_4$ (see Figure~\ref{subfig:Rolling_forecast_x4}), indicating that the \ac{DDM} wrongly predict anode effects in these cases. Moreover, the voltage drop due to electrical resistance is given by $\frac{u_2 u_5}{2620 g_2}$, where $u_2$ is the line current, $u_5$ is the \ac{ACD}, $2620 [m^2]$ is the total surface of the anodes and $g_2$ is the electrical conductivity. Within reasonable operational conditions, $\frac{1}{g_2}$ can be approximated as a function that increases linearly with the increasing mass ratio of alumina $c_{x_2}$. The modeling error in $x_4$ can therefore propagate to $x_6$. After approximately eight hours, the error bound of \ac{CoSTA} models shows that one of the \ac{CoSTA} models calculates an instantaneous overestimate of $x_6$, followed by an instantaneously underestimate of $x_6$. A possible explanation is that the \ac{CoSTA} model first erroneously predicts the anode effect. The underestimate of $x_6$ that instantaneously follows can possibly be caused by an underestimate of $c_{x_2}$ that is lower than $c_{x_2,crit}$ which leads to negative $P_{el}$ values in the model. 

\paragraph{Temperature in the side ledge $x_7$:}

The change of temperature in the side ledge $\dot{x}_7$ is determined by the heat balance in the side ledge. That is, the heat transfer from the bath to the side ledge $Q_{liq-sl}$, the heat transfer from the side ledge to the side wall $Q_{sl-wall}$, and the energy $E_{tc, sol}$ required to heat frozen side ledge to liquidus temperature from side ledge temperature. The change of side ledge temperature depends on the side ledge thickness $x_1$, the bath temperature $x_6$, the side ledge temperature $x_7$, the wall temperature $x_8$ and the liquidus temperature $g_1$. As argued above, for the \ac{PBM} modeling errors in $x_1, \; x_6, \; x_7, x_8$, and $g_1$ will propagate as modeling errors in the side ledge temperature change $\dot{x}_7$.  For the \ac{DDM} and \ac{CoSTA} models, the error bounds for the modeling errors of $x_7$ shown in Figure~\ref{subfig:Rolling_forecast_x7} are mostly growing simultaneously with error spikes in the error bound of $x_6$, presumably caused by erroneously predicted anode effects, as explained above.  

\paragraph{Temperature in the wall $x_8$:}
Figure~\ref{fig:Rolling_forecast}h shows that
The temperature of the side wall $x_8$ is changing according to the heat transfer from the side ledge to the wall $Q_{sl-wall}$, and the heat transfer from the wall to the ambient $Q_{wall-0}$. Changes in the wall temperature $\dot{x_8}$ depend on the side ledge temperature $x_7$, the wall temperature $x_8$, and the side ledge thickness $x_1$. \ac{PBM} modeling errors of these states at time $k$ propagate as modeling errors in the side wall temperature $x_8$ in the next time step, $k+1$. Hence, the \ac{PBM} will, with correct inputs always model the correct $\dot{x}_8$ since the \ac{PBM} model of $\dot{x}_8$ is equal to the simulator.

\section{Conclusions and future work}
\label{sec:conclusions}

In this work, we presented a recently developed approach in modeling called the Corrective Source Term Approach (CoSTA). CoSTA belongs to a family of hybrid analysis and modeling (HAM) tools where physical-based models (PBM) and data-driven models (DDM) are combined to exploit the best of both approaches while eliminating their weaknesses. The method was applied to model an aluminum extraction process governed by very complex physics. First, a detailed high-fidelity simulator was used to generate a dataset treated as the ground truth. Then, an ablated model was created by setting an internal variable of the simulator to a constant. Finally, the ablated model was supplemented with a corrective source term modeled using a \ac{NN} that compensated for the ignored physics. The main conclusions from the study are as follows: 
\begin{itemize}
    \item CoSTA, in all the scenarios investigated, could correct for the ignored physics and hence was consistently more accurate than both the PBM and DDM over a reasonably long time horizon. 
    \item CoSTA was consistently more stable and consistent in predictions when compared to pure DDM. 
    \item Regularizing the networks using $\ell_1$ weight decay was found to be effective in improving model stability in both \ac{DDM} and \ac{CoSTA}.
\end{itemize}

One significant benefit of the CoSTA approach is that it can maximize the utilization of domain knowledge, leading to reliance on black-box \ac{DDM} for modeling only those physics that are either not known or are poorly known.
Although it remains to be investigated in future work, it can be expected that much simpler models will be sufficient for modeling the corrective source terms.
These source terms can then be investigated to achieve additional insight giving more confidence in the model. 
Even if it is not possible to interpret the source terms, it should still be possible to place bounds on their outputs using domain knowledge. 
This can be used as an inbuilt sanity check mechanism in the system.
For example, since we know the amount of energy put into the system, the source terms for the energy equation will be bounded, so any \ac{NN}-generated source term violating this bound can be confidently rejected, making the models more attractive for high-stake applications like the one considered here. 
Another topic that would be worth investigating will be the robustness of the method to noise. 

\section*{Acknowledgments}
This work was supported by the project TAPI: Towards Autonomy in Process Industries (grant no. 294544), and EXAIGON: Explainable AI systems for gradual industry adoption (grant no. 304843)

\section*{Author contributions}
\textit{\textbf{Haakon Robinson}}: Methodology, Investigation, Software, Visualization, Writing (Original Draft) \textit{\textbf{Erlend Lundby}}: Methodology, Investigation, Software, Visualization, Writing (Original Draft) \textit{\textbf{Adil Rasheed}}: Conceptualization, Writing (Review \& Editing), Supervision, Funding acquisition \textit{\textbf{Jan Tommy Gravdahl}}: Supervision, Writing (Review \& Editing), Funding acquisition. First and second authors made equal contributions. 

\bibliographystyle{elsarticle-harv} 
\bibliography{ref.bib}
\end{document}